\documentclass[10pt]{article} 

\usepackage[preprint]{rlj} 

\usepackage{amssymb}            
\usepackage{mathtools}          
\usepackage{mathrsfs}           
\usepackage{graphicx}           
\usepackage{subcaption}         
\usepackage[space]{grffile}     
\usepackage{url}                
\usepackage{lipsum}             
\usepackage{algorithm, algorithmic}
\usepackage{wrapfig}

\newcommand{\State}[0]{\mathcal{S}}
\newcommand{\Action}[0]{\mathcal{A}}
\newcommand{\Reward}[0]{\mathcal{R}}

\newcommand{\Transition}[0]{\mathcal{P}}
\newcommand{\Dataset}[0]{\mathcal{D}}
\newcommand{\dataset}[0]{\mathcal{D}}
\newcommand{\R}[0]{\mathbb{R}}

\title{Fine-Tuning without Performance Degradation}

\setrunningtitle{Fine-Tuning without Performance Degradation}

\author{Han Wang\textsuperscript{1,2}, Adam White\textsuperscript{1,2,3}, Martha White\textsuperscript{1,2,3}, 
}

\emails{\{han8, amw8, whitem\}@ualberta.ca}

\affiliations{
$^{1}$\textbf{Department of Computing Science, University of Alberta, Canada}\\
$^{2}$\textbf{Alberta Machine Intelligence Institute (Amii)}\\
$^{3}$\textbf{Canada CIFAR AI Chair}\\

}

\contribution{
We empirically demonstrate that existing algorithms either fail to increase performance during fine-tuning or exhibit significant performance degradation. Specifically, we show that several algorithms 
, including InAC, IQL and AWAC, largely prevent performance degradation but also learn very slowly during fintuing. Other algorithms, like PEX, Proto, SAC and SAC with Ensemble CQL, do not prevent performance degradation. 
}
{
Previous work showed some of these methods prevented performance degradation, but our more extensive results show that is not the case for all the offline datasets and environments we considered. To the best of our knowledge, no one has shown that InAC, IQL and AWAC prevent performance degradation, although this is potentially because they do not really provide an effective approach to fine-tuning.
}

\contribution{
We propose a practical fine-tuning algorithm by modifying an existing algorithm called Jump Start Reinforcement Learning (JSRL), introduced to improve fine-tuning performance. Our Automatic Jump Start algorithm is simple, using InAC for offline training combined with an automatic way to slowly move from InAC to SAC during fine-tuning. It combines the benefits of these base algorithms (minimal performance degradation from InAC and fast fine-tuning from SAC), and automatically balances between them without introducing new hyperparameters. 
}
{
JSRL \citep{jump2022uchendu} was introduced to leverage a given guide policy (potentially computed from offline data or a simulator), to bring or guide the learning exploration agent to different parts of the environment (not necessarily for fine-tuning). We found JSRL, as proposed, was not effective in our setting, but that the natural extension to incorporate InAC and SAC was effective, given a threshold parameter is well-tuned. We highlighted the sensitivity to this threshold parameter, which is used to decide how to slowly hand off control from the guide policy to the exploration policy. Automatic Jump Start eliminates the need for this threshold. Overall, our Automatic Jump Start algorithm is much more effective for our fine-tuning setting than the original JSRL algorithm, particularly towards real-world deployment where it is key to avoid performance degradation and hyperparameter tuning is restricted or even not feasible. 
}


\summary{
    Fine-tuning policies learned offline remains a major challenge in application domains.
    Monotonic performance improvement during \emph{fine-tuning} is often challenging, as agents typically experience performance degradation at the early fine-tuning stage.
    The community has identified multiple difficulties in fine-tuning a learned network online,
    however, the majority of progress has focused on improving learning efficiency during fine-tuning. In practice, this comes at a serious cost during fine-tuning: initially, agent performance degrades as the agent explores and effectively overrides the policy learned offline. We show across a range of settings, many offline-to-online algorithms exhibit either (1) performance degradation or (2) slow learning (sometimes effectively no improvement) during fine-tuning. We introduce a new fine-tuning algorithm, based on an algorithm called Jump Start, that gradually allows more exploration based on online estimates of performance. Empirically, this approach achieves fast fine-tuning and significantly reduces performance degradations compared with existing algorithms designed to do the same.  
}

\begin{document}

\maketitle  

\begin{abstract}
    Fine-tuning policies learned offline remains a major challenge in application domains.
    Monotonic performance improvement during \emph{fine-tuning} is often challenging, as agents typically experience performance degradation at the early fine-tuning stage.
    The community has identified multiple difficulties in fine-tuning a learned network online,
    however, the majority of progress has focused on improving learning efficiency during fine-tuning. In practice, this comes at a serious cost during fine-tuning: initially, agent performance degrades as the agent explores and effectively overrides the policy learned offline. We show across a range of settings, many offline-to-online algorithms exhibit either (1) performance degradation or (2) slow learning (sometimes effectively no improvement) during fine-tuning. We introduce a new fine-tuning algorithm, based on an algorithm called Jump Start, that gradually allows more exploration based on online estimates of performance. Empirically, this approach achieves fast fine-tuning and significantly reduces performance degradations compared with existing algorithms designed to do the same.
\end{abstract}

\section{Introduction}
\label{intro}

{Fine-tuning} allows policies learned offline to improve with additional interaction in the real environment. The agent begins with a policy learned offline and continuously adjusts it through interaction with the environment. Intuitively, starting with a learned policy provides a warm start, allowing the agent to learn more efficiently than learning from scratch, requiring fewer samples to achieve a certain level of performance. In addition, fine-tuning should also enable the agent to improve on a suboptimal initial policy, which often occurs if the offline dataset has low coverage or was generated by a suboptimal policy like a human operator. 

Monotonic policy improvement, however, remains challenging, as performance degradation is often observed in the early fine-tuning stage. In low-risk scenarios, it may be acceptable for performance to degrade before it gets better; however, in many cases, sharp performance degradation is unacceptable. Consider controlling the heating and cooling system in a hospital. The behavior policy used for dataset collection often represents current control strategies, such as heating and cooling a hospital. This behavior is likely not optimal regarding energy efficiency, but reasonable in terms of comfort and temperature ranges. It would not be acceptable for the agent to make the hospital uncomfortable for patients and staff; the agent needs to maintain the same level of comfort and slowly improve efficiency.

Various hypotheses have been explored in the literature to understand this performance degradation. We classify them into three main categories. \textbf{Representation collapse and catastrophic forgetting} is the idea that fine-tuning alters the network's hidden layers \citep{peters-etal-2019-tune, merchant-etal-2020-happens, zhang2021revisiting, zhou-srikumar-2022-closer} degrading the agent's learned representation and ultimately the policy. 
\citep{razdaibiedina2022improving, aghajanyan2021better,campos2021finetuning,zhang2023policy,song2023hybrid}. Several approaches have been introduced to constrain the update on weights \citep{li2023proto, luo2023finetuning}, or ensure the offline-trained representation generalizes across multiple tasks \citep{razdaibiedina2022improving}.
In \textbf{distribution shift}, data collected from online interactions has a different distribution from the offline data, leading to severe bootstrap errors~\citep{lee2021offlinetoonline, zhao2022adaptive}. There have been attempts to stabilize training by incorporating offline data with online data to control the data distribution \citep{lee2021offlinetoonline, ball2023efficient}.
Finally, the action values have \textbf{overestimation} during offline learning which interacts negatively with bootstrapping \citep{lee2021offlinetoonline, nakamoto2023calql}.
Ensemble networks and Conservative Q-Learning (CQL) have been introduced to mitigate this value inflation~\citep{lee2021offlinetoonline, nakamoto2023calql, zhao2023improving,kumar2020conservative}. In practice, the methods discussed above either result in little improvement during fine-tuning, because they are too conservative, or exhibit sudden, dramatic performance degradation, as we later show.

Recently, a new way of balancing conservatism and exploiting new experiences was introduced, but its performance critically depends on tuning several key hyperparameters. The idea of the Jump-Start algorithm is simple~\citep{jump2022uchendu}. Deviation from the offline policy in fine-tuning near the start-state distribution is risky because an exploration step may put the agent in a new region of the state space where the offline policies perform poorly for the rest of the trajectory, which in turn could induce representation collapse. Conversely, deviating near the end of the trajectory is likely to have little impact on the policy overall. The agent can slowly work backwards from the end of the trajectory, each time following the fine-tuned policy more and more. This approach has been shown to work well across D4RL problems, but the big question is how quickly to step back and when. Currently, the algorithm relies on hyperparameter sweeps in the true environment, but this is unrealistic. The whole purpose of offline RL is to find a policy that works well when it is deployed (with fine-tuning), without assuming access to the deployment environment. Going back to our hospital settings, it is like assuming we can conduct hyperparameter sweeps of the algorithm controlling the HVAC system while patients and doctors are in the building!

In this paper, we focus on the practical aspects of performance degradation during the fine-tuning phase of offline-to-online RL. First, we empirically demonstrate that existing algorithms either fail to increase performance during fine-tuning or exhibit dramatic performance degradation. Soft Actor Critic, for example, when used in both offline and fine-tuning phases, induces a massive performance degradation but also achieves the largest improvement during fine-tuning. We show that the Jump-Start algorithm can dramatically reduce performance degradation and achieve good fine-tuning (less so than SAC), but its hyperparameters must be tuned during fine-tuning for each environment and data collection policy. We introduce a new Jump-Start algorithm to eliminate the need for hyperparameter tuning. Our Automatic Jump-Start algorithm maintains an offline policy and an exploration policy fine-tuned from recent online experience. On each step, the algorithm decides whether to follow the offline policy or the exploration policy based on an online estimate of performance. Empirically we find that Automatic Jump-Start is comparable to Jump-Start with hyperparameter tuning, sensibly increasing the amount of actions from the fine-tuned policy. Our new algorithm and set of empirical results represent a small but significant step towards reducing hyperparameter tuning for real-world offline-to-online RL.

\section{Background and Problem Setting} \label{sec:problem}

In this paper we consider problems formulated as Markov Decision Processes (MDP), where $\mathcal{M}=\left<\State, \Action, \Transition, \Reward, \gamma \right>$. $\State \in \R^d$ represents the state space, $\Action\in \R^k$ is the action space, and the transition function $\Transition: \State \times \Action \times \State \rightarrow [0, 1]$ describes the probability of transitioning from a state action pair to another state.
The reward function is defined as $\Reward: \State \times \Action \times \State \rightarrow \mathbb{R}$ and a discount factor is $\gamma \in [0, 1)$, which is zero at the terminal state \citep{white2017unifying}. The goal is to continually improve the agent's policy, $\pi: \State \times \Action \rightarrow [0, 1]$, to maximize the discounted sum of the future reward called the {\em return}, $G_t \doteq r_{t+1} + \gamma r_{t+2} + \gamma^2 r_{t+3} + \dots$.

All the methods we consider in this paper learn a state-action value function. The agent maintains an action value estimation function $Q_\theta$, parameterized by $\theta\in\mathbb{R}^d$, to estimate the expected return under $\pi$: $\mathbb{E}[G_t|S_t=s, A_t=a]$, starting from state-action pair $<s,a>$ and taking actions according to $\pi$. An agent may also learn a state value function $V_\psi(s)\doteq\sum_a\pi(a|s)Q_\theta(s,a)$.

The offline-to-online learning problem consists of two phases: offline learning, followed by fine-tuning. In the offline learning phase, a policy, $\pi_\phi: \State \times \Action \rightarrow [0, 1]$ parameterized by $\phi$, is learned from a static dataset $\Dataset = \{\left<s, a, s', r, \gamma \right>_i\}$, generated by some (potentially unknown) policy $\pi_\Dataset$. In the online, fine-tuning phase that follows, the agent interacts with the environment---the same environment the original data, $\Dataset$ was collected in---with the objective of further improving $\pi_\phi$.

We define policy degradation if the fine-tuning phase decreases the performance of $\pi_\phi$. Precisely, the policy learned offline $\pi_\phi$ has performance $p_0$, estimated by rolling out the fixed policy in the environment. During fine-tuning, the agent collects online returns $p_1, p_2, ...$. We take the worst performance and measure the degradation as $(\min(p_1, p_2, ...) - p_0) / p_0$.

Multiple algorithms have been designed specifically for this offline-online setting, such as AWAC \citep{nair2021awac}, PROTO \citep{li2023proto}, Off2On \cite{lee2021offlinetoonline}, Adaptive Behavior Cloning \cite{zhao2022adaptive}, and Policy Expansion \citep{zhang2023policy}. Alternatively, offline learning can be done with an offline reinforcement learning algorithm such as CQL \citep{kumar2020conservative}, IQL \citep{kostrikov2022offline}, or InAC \citep{xiao2023the}. Then, one may use an algorithm like SAC \citep{haarnoja2018soft} to fine-tune the policy learned offline.


Many of the algorithms we explore in this paper are based on SAC and InAC.
SAC minimizes a KL-divergence to the Boltzmann policy $\pi(a|s) \propto \exp \frac{q(s,a)}{\tau}$, where $\tau$ is the entropy parameter. A higher $\tau$ encourages higher entropy and stronger exploration.
InAC is an offline learning algorithm with an update similar to SAC but designed to prevent bootstrapping from out-of-distribution actions \citep{xiao2023the}.  InAC constrains the actor update with by using a slightly different target policy in the KL-divergence 
\begin{align*}
    \pi(a|s) \propto \exp \left(\frac{q(s,a)}{\tau} \right)
    = \frac{\pi_\dataset(a|s)}{\pi_\dataset(a|s)} \exp \left(\frac{q(s,a)}{\tau} \right)
    = \pi_\dataset(a|s) \exp \left(\frac{q(s,a)}{\tau} - \log\pi_\dataset(a|s) \right) ,
    \label{eq:inac-optimal-policy}
\end{align*}
where $0 \cdot \infty = 0$. When $\pi_\dataset(a|s)=0$---the action is out-of-distribution---the target policy similarily has $\pi(a|s) = 0$. Otherwise, for actions with $\pi_\dataset(a|s)>0$, the distribution is the same as SAC.

\section{Performance Degradation and Why it happens}\label{sec:perf_sac}

In this section, we identify and investigate the occurrence of severe performance degradation during fine-tuning. We first highlight performance degradation in the early stages of fine-tuning and analyze the impact of different initializations. Then, we investigate how the policy changes in early fine-tuning and discuss the challenges introduced by exploration. We use D4RL datasets~\citep{fu2020d4rl} for all the experiments that follow. 

\subsection{The Existence of Performance Degradation}

\begin{wrapfigure}[18]{11}{.5\textwidth}
    \centering
    \vspace{-0.6cm}
    \includegraphics[width=.5\textwidth]{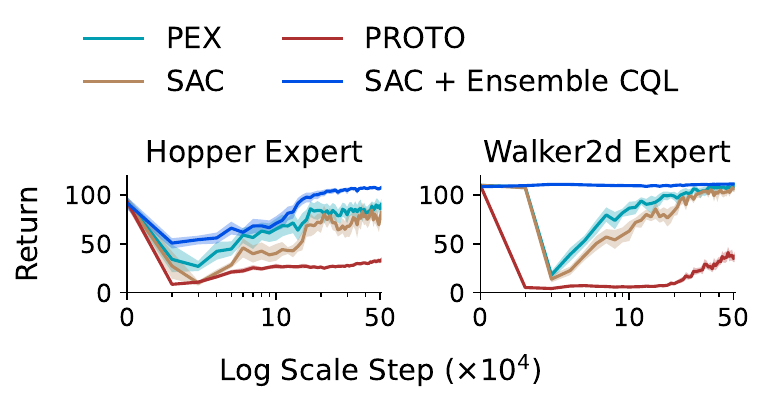}
    \captionof{figure}{\label{fig:expert_drop}{All algorithms tested could not prevent performance degradation when fine-tuning a near-optimal policy. The x-axis represents the timestep ($\times 10^4$) during fine-tuning, while the y-axis is the normalized return averaged over 15 runs. The shaded area indicates the 95\% bootstrap confidence interval. SAC fine-tuning a the offline InAC policy was included as a baseline. SAC+Ensemble CQL used 10 networks.  }
    }
\end{wrapfigure}

We evaluated several existing algorithms initialized with a near-optimal policy learned from the Expert dataset, using the original hyperparameters reported in their respective papers.  The main question we seek to answer here is which algorithms exhibit performance drop and why?
We choose to include representative algorithms from the following categories: preserving a fixed offline policy, constraining policy change during fine-tuning, and ensemble methods. Each of these classes have been show to provide some protection against policy degradation. 
Policy Expansion (PEX) saves a fixed copy of the policy learned offline. During fine-tuning, PEX updates a separate policy which is initialized to the offline policy. PEX samples actions probabilistically from both policies according to the learned value estimates for each \citep{zhang2023policy}.
PROTO fine tunes the policy learned offline in a conservative way via a KL penalty \citep{li2023proto}.  
We also include a method that combines ensemble CQL updates offline and then during fine-tuning the updates for the ensembles is switched to SAC instead of CQL~\citep{zhao2023improving}. This method is both conservative during offline learning and attempts to leverage the benefits of ensembles. The ensemble size used was the same as previous work~\citep{zhao2023improving}. We used InAC during offline training for PEX and PROTO because this method was been previously shown to be effective in offline training~\citep{xiao2023the}. 

Figure \ref{fig:expert_drop} summarizes the results. We report normalized performance using the return bounds provided in D4RL. Each algorithm experienced performance degradation in at least one of the two environments and no algorithm fully prevented severe performance degradation when the offline policy is learned from high quality data. We include additional analysis of the failure of ensemble methods and conservatism estimation in supplementary material \ref{apdx:ensemble}.

\begin{figure}[hb]
    \centering
    \includegraphics[width=0.85\textwidth]{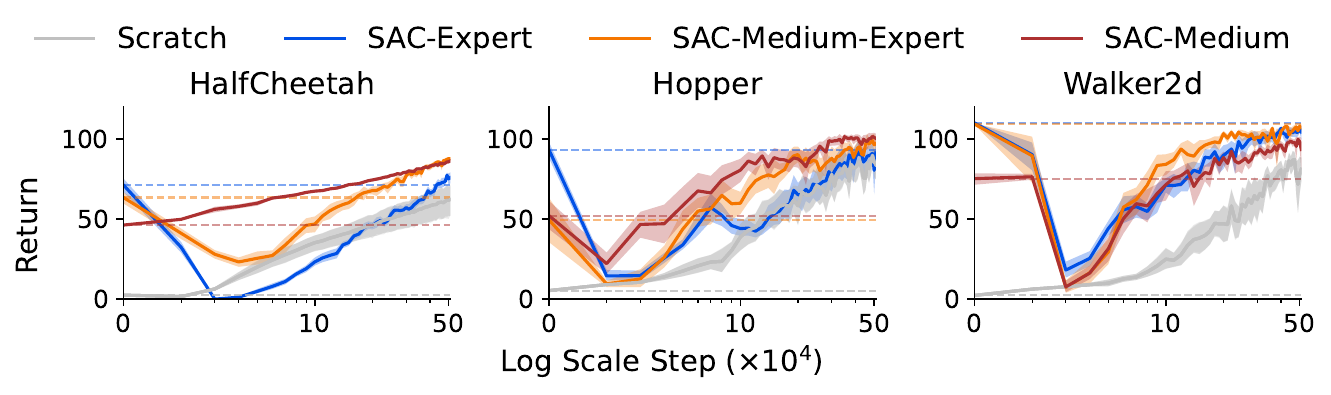}
    \vspace{-0.3cm}
    \caption{\label{fig:ft_behavior_policy}{Performance degredation of SAC is related to the quality of the policy used to generate the data for offline training. In all cases, fine-tuned SAC eventually outperforms a the simple baseline of using SAC without offline training (labelled Scratch). The y-axis reports normalized performance averaged over 15 runs, while the x-axis denotes the time step $(\times 10^4)$ during fine-tuning. The shaded area indicates the 95\% bootstrap confidence interval. The horizontal dashed line indicates the performance at initialization for each dataset (color matched). }
    }
\end{figure}

Perhaps the extent of performance degradation is related to quality of the learned offline policy. To test this hypothesis, we trained a policy offline with InAC using three different datasets of different qualities and then used SAC for fine-tuning. We did this in three different Mujoco tasks. Figure \ref{fig:ft_behavior_policy} contains the results. In HalfCheetach, we see that agents initialized with better policies suffer large performance degradation during fine-tuning. In the other two environments, all agents dropped to a near-random level performance before eventually recovering. It appears that agents experience greater performance degradation when the initial performance was better. 

\subsection{Encountering New States and Actions During Fine-tuning is Problematic}\label{sec:overestimation}

\begin{figure}[t]
    \centering
    \vspace{-0.3cm}
    \includegraphics[width=0.7\textwidth]{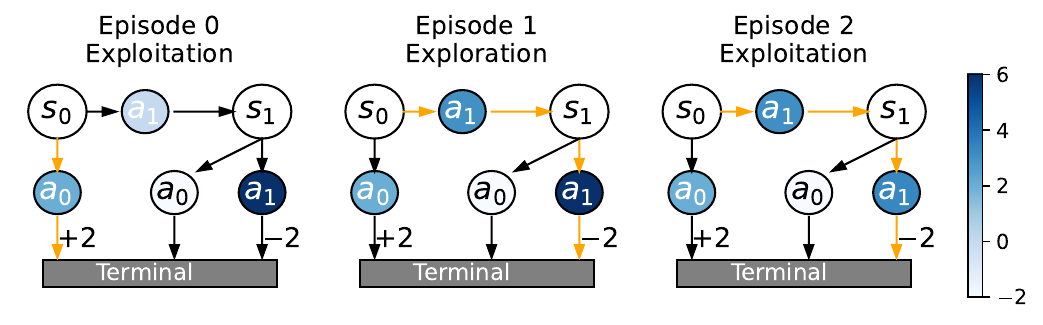}
    \vspace{-0.3cm}
    \caption{\label{fig:oe_example}{
    The effect of overestimation in states that are not covered by the dataset. 
    Each subplot visualizes the state-action pairs in the environment at different time steps. The subtitle specifies whether the agent exploited at state $s_0$. Action values are represented by color intensity, with deeper colors indicating higher estimates. Yellow arrows refer to the path taken by the agent in that episode.}
    }
\end{figure}

A likely culprit for performance degradation is overestimation of unseen states. Conservative value estimates, given by algorithms like CQL and InAC, are only conservative for states in the dataset. Once online, however, the agent is likely to encounter new states and be potentially skewed by these overestimates. 

To better understand this, let us consider a toy example in \ref{fig:oe_example}. 
Imagine in offline learning that $s_1$ is not in the dataset and we happen to have an overestimate for $(s_1, a_1)$ (indicated by the dark blue), potentially due to an arbitrary initialization of the values for that unobserved state. Offline training does not change this arbitrary initial value, because the values for $s_1$ are never updated. During offline training, the agent learns that $(s_0, a_0)$ has a reasonably high value, which is in fact the optimal choice. Once deployed, it will correctly take action $a_0$ from $s_0$ until it explores and takes action $a_1$. Once seeing state $s_1$, the agent will update the value of $(s_0, a_1)$ by bootstrapping from the artificially high value $(s_1, a_1)$. Now the agent will incorrectly think $(s_0,a_1)$ is high-value, it will start going this suboptimal path, performance will degrade and it will take time for the agent to adjust those action-value estimates. Ultimately, it will relearn that $a_0$ is optimal from $s_0$, but in the interim we will see exactly the performance degradation we say above.


\begin{figure}[t]
    \centering
    \vspace{-0.3cm}
    \subfloat[\scriptsize InAC, HC]{\includegraphics[clip, trim={0.5cm 0.4cm 0.1cm 0 }, width=0.16\textwidth]{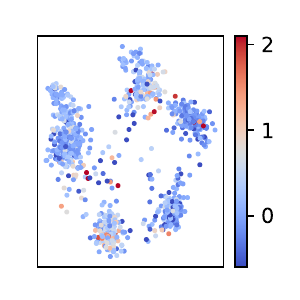}\label{oe_halfcheetah_inac}}
    \subfloat[\scriptsize SAC, HC]{\includegraphics[clip, trim={0.5cm 0.4cm 0.1cm 0 }, width=0.16\textwidth]{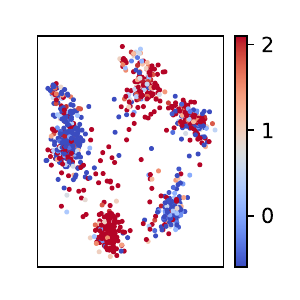}\label{oe_halfcheetah_sac}}
    \subfloat[\scriptsize InAC, Hopper]{\includegraphics[clip, trim={0.5cm 0.4cm 0.1cm 0 }, width=0.16\textwidth]{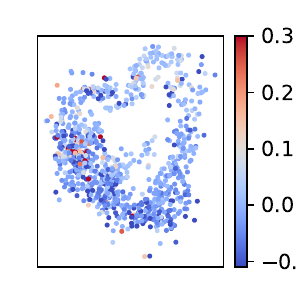}\label{oe_hopper_inac}}
    \subfloat[\scriptsize SAC, Hopper]{\includegraphics[clip, trim={0.5cm 0.4cm 0.1cm 0 }, width=0.16\textwidth]{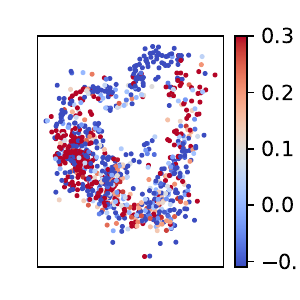}\label{oe_hopper_sac}}
    \subfloat[\scriptsize InAC, Walker2D]{\includegraphics[clip, trim={0.5cm 0.4cm 0.1cm 0 }, width=0.16\textwidth]{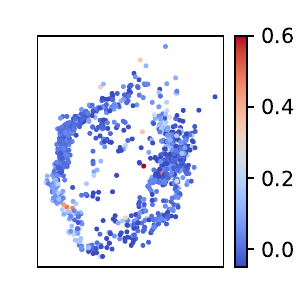}\label{oe_walker_inac}}
    \subfloat[\scriptsize SAC, Walker2D]{\includegraphics[clip, trim={0.5cm 0.4cm 0.1cm 0 }, width=0.16\textwidth]{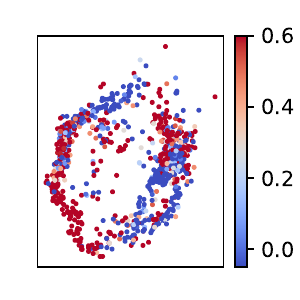}\label{oe_walker_sac}}

    \caption{\label{fig:oe_d4rl}{After 10,000 updates, SAC shifted toward selecting actions with higher initial value estimates. Each scatter represents a randomly sampled state from the offline dataset used for policy learning, with coordinates computed via PCA. The color of each scatter indicates the value difference between the updated policy and the initialized policy, as measured by the initial critic. A more intense red indicates a larger difference, meaning fine-tuning shifts the policy toward actions with higher initial value estimates. We set the color bar in the same environment to the same value range for clear visualization and shorten HalfCheetah to HC.}}
\end{figure}





We also checked if SAC was directly suffering from overestimation for out-of-distribution actions, not just bootstrapping off of values for unvisited states. Let $q_0$ be the action-values learned offline by InAC. We check how much SAC shifts action selection towards actions with high-value under $q_0$ for states in the offline dataset, but where InAC does not prefer those actions. If they are high-valued but not selected by InAC, it indicaes they are out-of-distribution actions. We report this in Figure \ref{fig:oe_d4rl}, where the more red there is in the scatterplot for SAC the more likely that it shifted to these out-of-distribution actions with arbitrarily high-value. This plot is generated by letting SAC fine-tune for 10k steps, which is generally a point where its performance has degraded, and then checking the value of SAC's policy according to $q_0$. Specifically, we subtract the value of SAC's policy under $q_0$ from the value of the offline policy under $q_0$. We provided InAC as a baseline, to show that unlike SAC, it generally did not shift value to these out-of-distribution actions. 



While visiting the unseen states can be problematic during fine-tuning, fully preventing this process is undesirable. Similar to exploration in pure online learning, visiting unseen states and actions can help the agent search for a better policy. We did also explore the use of ensembles to provide conservative values for unseen states, but ran into exactly this problem: it could prevent degradation but also often prevented learning during fine-tuning (see Supplement \ref{apdx:ensemble}). In the following sections, we investigate other approaches to prevent performance degradation but still promote fast learning.

\section{Offline Algorithms Mitigate Performance Degradation But Learn Slowly}\label{sec:inac_ft}

In this section we show that several (conservative) offline algorithms can prevent performance degradation, but that they also learn too slowly. They provide an inadequate solution to the performance degradation issue, because sample efficiency online is so severely sacrificed. These results motivate designing an algorithm that strikes a balance, as we do in the following section, by building on the algorithms in this section.

We test several offline algorithms, that can continue to update online. These include InAC, Implicit Q Learning (IQL) \citep{kostrikov2022offline} and AWAC \citep{nair2021awac}.  
For InAC, we keep the same entropy and learning rate used in offline learning without additional tuning. For IQL and AWAC, we used the same entropy and learning rate as reported in the original paper. We include SAC as a baseline, where it's initial policy is trained offline using InAC, as in the previous section.

\begin{figure}[ht]
    \centering
    \vspace{-0.3cm}
    \includegraphics[width=0.8\textwidth]{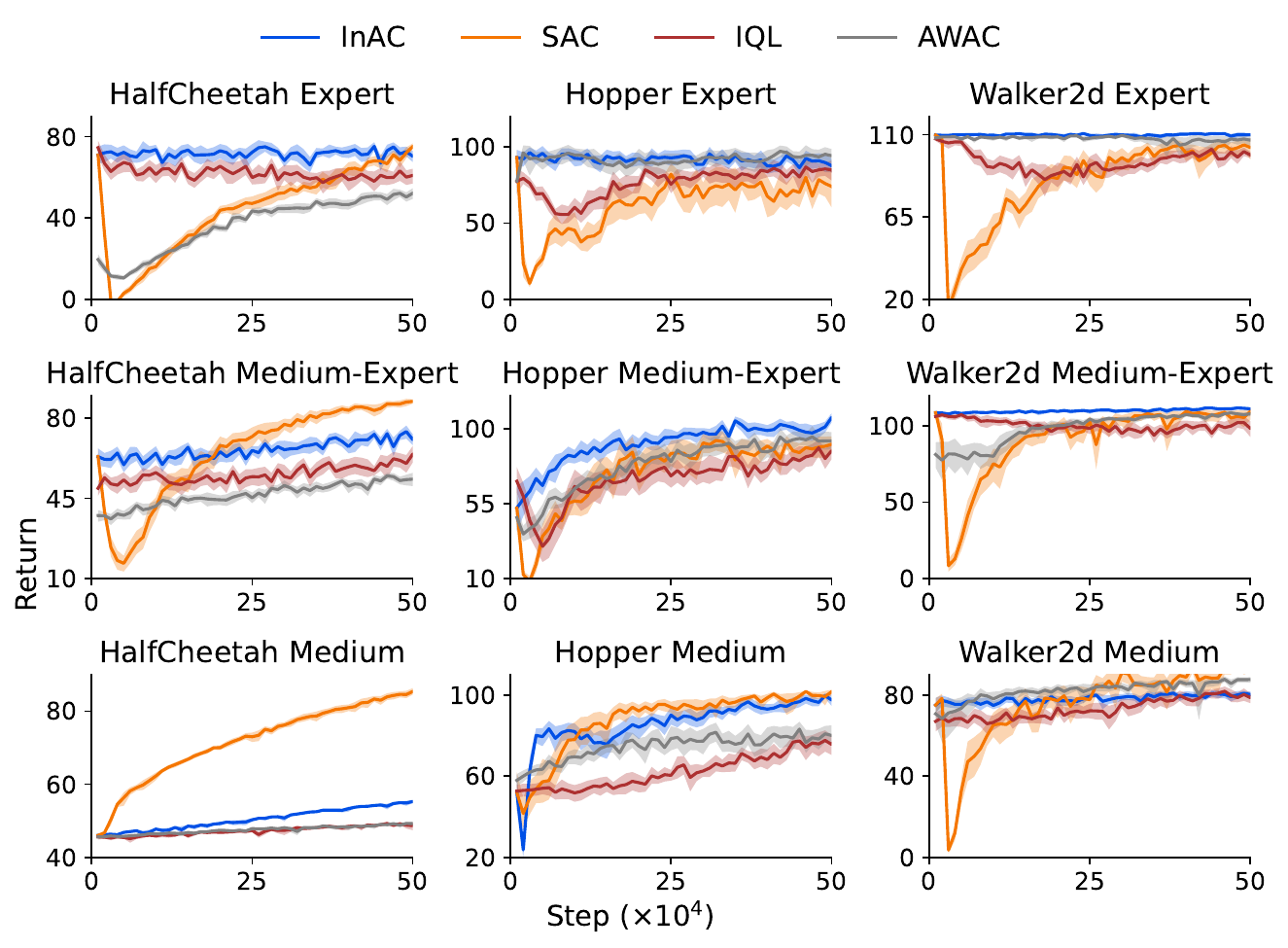}
    \vspace{-0.3cm}
    \caption{\label{fig:compare_inac_sac}{InAC demonstrated more stable performance when the policy initialization was near-optimal, but underperformed SAC when the initialization was suboptimal. Each row presents learning curves across different environments, while each column corresponds to a different policy initialization, trained on the Expert, Medium-Expert, and Medium datasets, respectively. The x-axis is the time step $(\times 10^4)$, and the y-axis is the normalized return. The shaded area refers to the $95\%$ confidence interval.}
    }
\end{figure}

We can see that all three methods largely prevent performance degradation, though IQL faired more poorly in this regard. In
Figure \ref{fig:compare_inac_sac}, we can see that InAC generally performs the best, avoiding performance drop but also allowing for the most improvement during fine-tuning. All the methods are significantly outperformed by SAC in a couple of environments, but at the same time SAC suffers from significant performance degradation.  

\section{Combining the Benefits of InAC and SAC through Controlled Exploration}\label{sec:ajs}

Achieving a better policy often requires exploring new regions of the state-action space, which may lead to temporary suboptimal performance. A natural assumption is that some degree of performance degradation is inevitable.
If a linear path exists from the initial policy to a better policy that the fine-tuned agent converges to, performance may monotonically improve. In practice, however, it is uncommon to see simple linear paths between such policies (detailed discussion is provided in supplementary materials \ref{sec:linear_comb}).
Although small degradations may be expected during learning, the extent of performance degradation we observe is excessive. In this section, we leverage the idea behind Jump Start RL \citep{jump2022uchendu}, which was introduced to improve exploration, to design an algorithm that largely avoids performance degradation like InAC but learns much more quickly during fine-tuning like SAC.


\subsection{Hyperparameter Sensitivity for a Vanilla Variant of Jump Start}

In this section we discuss issues with a naive extension of Jump Start RL (JSRL) to our setting. JSRL was introduced to improve exploration, by supplementing with a policy learned offline or in a simulator. The purpose was to make learning faster, rather than prevent performance degradation. However, the idea behind JSRL when combined with InAC and  SAC is a promising path to avoid performance degradation and promote sample efficiency.  

JSRL maintains two policies: a fixed \emph{guide policy} $\pi^g$ that rolls out in the environment and controls the starting point of exploration, and an online learnable \emph{exploration policy} $\pi^e$, initialized to $\pi^g$ and updated online. The agent follows the guide policy during the first $h$ steps and then switches to $\pi^e$. It starts learning with $h = H-1$, the horizon of the episodic problem, meaning it starts by only running $\pi^g$ for most of the episode and allowing $\pi^e$ to take an action only on the last step.
The agent monitors its performance using a sliding window on the latest returns. The guide step $h$ reduces only when the latest performance outperforms the previous best return by a tolerance $\epsilon$. We can use JSRL with InAC to learn $\pi^g$ and SAC to update $\pi^e$ online.

\begin{figure*}[h]
    \centering
    \includegraphics[width=0.8\textwidth]{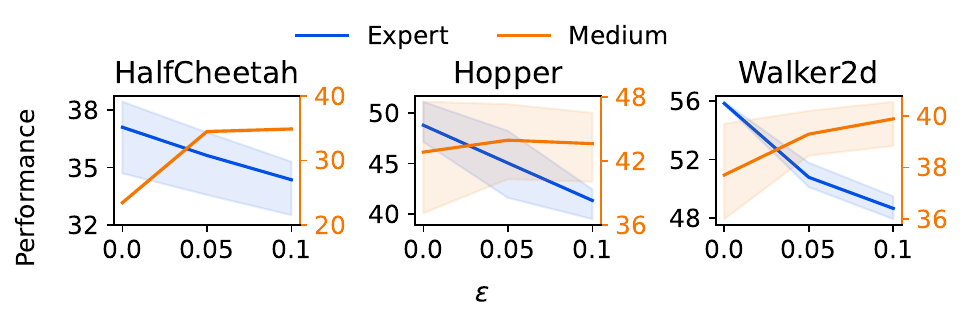}
    \caption{\label{fig:js_sensitivity}{When the policy initialization was near-optimal, a smaller $\epsilon$ in JSRL was preferred. Conversely, a larger $\epsilon$ worked better when the initialization was suboptimal. The plot reports sensitivity curves for policies learned from the Expert and Medium datasets separately. The x-axis represents the performance threshold of JSRL, $\epsilon$, and the y-axis is the area under-curve of the first 500,000 steps of fine-tuning. The expert curve uses the left y-axis, while the medium curve uses the right y-axis. The average was calculated over 5 seeds. The shaded area refers to $95\%$ bootstrap confidence interval.}}
\end{figure*}

Unfortunately, JSRL's performance can be sensitive to $\epsilon$.
A large tolerance $\epsilon$ leads to a rapidly decreasing guide step, encouraging aggressive exploration, which may result in performance degradation. In contrast, an overly small tolerance constrains exploration and limits policy improvement. Figure \ref{fig:js_sensitivity} illustrates how performance varied with different tolerances $\epsilon$, across three environments with expert vs medium offline datasets. The crossing lines highlight that very different parameters $\epsilon$ are needed for the Expert datasets versus the Medium datasets. 

\subsection{Automatic Jump Start}

In this section we propose Automatic Jump Start (AJS) to eliminate the performance threshold parameter in JSRL. 
The key to automatically adjusting the exploration step $h$ is to provide a reliable comparison between the offline learned policy and the fine-tuned policy. An average of returns provides an unbiased estimate of policy performance, but suffers from high variance, particularly when the window size is small. 
To mitigate this variance, we replace the moving window average with an off-policy estimation (OPE) method. A reliable estimator eliminates the need for a threshold $\epsilon$ and we find we can consistently set it to 0.

For OPE, we use Fitted Q evaluation (FQE) \citep{le2019fqe}. FQE estimates the action-values $Q_{\theta}$ for a given evaluation policy $\pi_e$, using a dataset of tuples $\{(s_i, a_i, s_{i}', r_{i})\}_{i=1}^n$ and Bellman targets $r_{i} + \gamma Q_{\theta} (s'_{i}, a')$, where $a' \sim \pi_e(s'_{i})$. The performance estimate is the sample average over all start-states $s \in \mathcal{D}_0$ in the data,  $\frac{1}{|\mathcal{D}_0|} \sum_{s_0 \in \mathcal{D}_0} \mathbb{E}_{\pi_e} Q_{\theta}(s_0, A)]$. For ASJ, the evaluated policy $\pi_e$ consists of InAC for the first $h$ steps of the episode followed by SAC (see Algorithm \ref{alg_jspolicy}). The FQE estimator is trained using the offline dataset for the final InAC policy. Once fine-tuning starts, the estimator is also fine-tuned, doing a few updates after each episode.

We make one other small modification, which is that allow the guide policy to be updated using the same algorithm as in offline learning (InAC). This modification allows the agent to take advantage of the policy improvement without concern for performance degradation. As section \ref{sec:inac_ft} suggests, InAC fine-tuning does not suffer from severe performance degradation, so introducing InAC updates to the guide policy should maintain stability in Jump Start.
We provide the pseudocode for the complete AJS algorithm in the supplement, in Algorithm \ref{alg_jsh}.

\section{Evaluating the Automatic Jump-Start Algorithm} \label{sec:exp}

We evaluated the performance of AJS on HalfCheetach, Hopper, and Walker2d, using policies learned from Expert, Medium-Expert, and Medium datasets from D4RL \citep{fu2020d4rl}.
We focused on the practical scenario where hyperparameter tuning is difficult or impossible, thus setting hyperparameters to defaults without any tuning. All agents used the same learning rate in offline learning and fine-tuning. We ran 15 seeds for each dataset and environment pair. 
We first evaluate the performance of AJS, compared to InAC, PEX and SAC, and also to variants of JSRL, to test if it does strike a balance between performance degradation and learning speed. Because AJS better controls exploration during fine-tuning, we also evaluated its robustness to the entropy parameter. And finally we looked more closely at how FQE change $h$ during fine-tuning in ASJ, to see if it is appropriately reducing $h$ overtime. 

\subsection{Balancing Stability and Improvement}

We first compared AJS to algorithms used earlier in this paper, that prevented performance degradation but learned slowly (InAC) and to those that learned faster but did not prevent performance degradation (PEX, SAC). PEX was particularly added as a baseline because it also retains a copy of the offline-learned policy to encourage stable performance \citep{zhang2023policy}. We measure the level of performance degradation in a run by taking the worst online return from the agent $G_{\text{worst}}$ and reporting relative performance to the offline policy $G_{0}$: $G_{\text{worst}} - G_0) / G_0$.
We averaged this performance degradation across environment and dataset pairs. We also measure the final improvement, by using the average of the returns over the last 10\% of fine-tuning $G_{\text{final}}$ and reporting relative performance to the offline policy $G_{0}$: $G_{\text{final}} - G_0) / G_0$. 

\begin{figure}[t]
    \centering
    \vspace{-0.3cm}
    \includegraphics[width=0.9\textwidth]{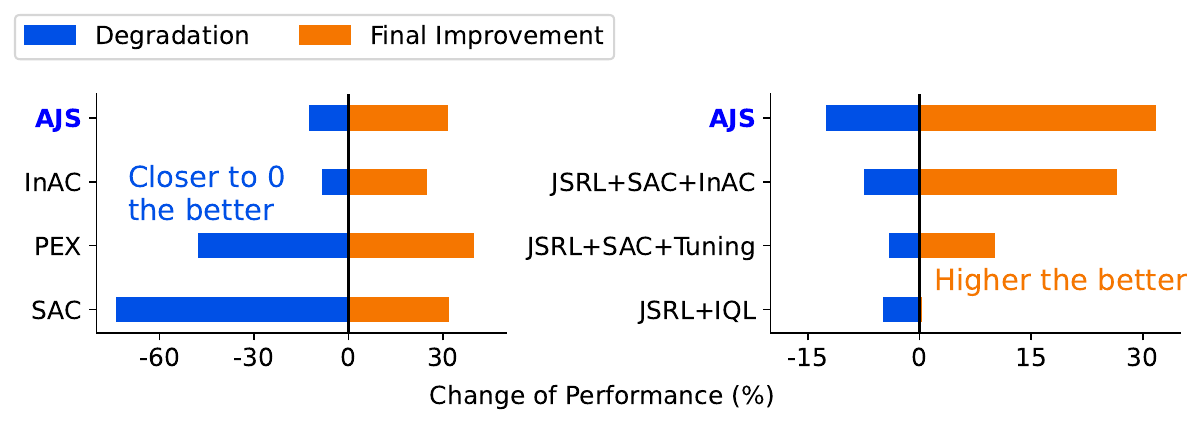}
    \caption{\label{fig:ajs_bar}{
                AJS balanced between stability and improvement with the default parameter setting. The experiment was conducted on HalfCheetah, Hopper, and Walker2d, using three different datasets from D4RL: Expert, Medium-Expert, and Medium. We tested 15 seeds for each of the 9 settings. The x-axis represents the average percentage of performance change. The y-axis indicates different agents. JSRL+SAC is omitted as it is highly similar to JSRL+IQL. The corresponding learning curves are in supplementary materials (Figure \ref{fig:ajs_curve}, \ref{fig:ajs_variants_curve}, and \ref{fig:ajs_tunedjs}).
            }}
\end{figure}

The results are reported in Figure \ref{fig:ajs_bar} (left). We can see that AJS has only a slightly higher performance degradation than InAC but also a higher final improvement. AJS has a similar final improvement to SAC, but significantly less performance degradation. 

We also compared AJS to several variants of JSRL. 
{JSRL+IQL} replicated the setting in the original paper. {JSRL+SAC} replaced IQL during fine-tuning by SAC. {JSRL+SAC+Tuning} uses SAC instead of IQL for fine-tuning, and we also tuned for the best performance tolerance in $\{0\%, 5\%, 10\%\}$. Otherwise, it performed poorly. All other JSRL variants used a default tolerance of 0. {JSRL+SAC+InAC} is the vanilla variant of JSRL closer, but where we also allowed the guide policy to be tuned by InAC. This variant is an ablation of the use of FQE in AJS to set $h$ because the only difference is that it relies on $\epsilon$ and windowed averages to set $h$ like in JSRL. All the JSRL variants used a window size of 5, as in the original paper. 

The results are reported in Figure \ref{fig:ajs_bar} (right). We can see that AJS has significantly higher final performance than the JSRL variants. With a tolerance of zero, these algorithms are slightly more conservative, meaning they have lower performance degradation but also slower improvement. As pointed out above, though, there is not a single tolerance $\epsilon$ that would work well across settings, and so we will see this trade-off. AJS provides a good balance, without having to consider tuning this hyperparameter. It is interesting to note that JSRL using InAC offline and SAC online performs notably better than the original version combined with IQL and notably better than the version where IQL is learned offline and SAC used online. 


\subsection{Robustness to the InAC Entropy}

\begin{wrapfigure}{R}{.5\textwidth}
\vspace{-0.5cm}
    \centering
    \includegraphics[width=0.5\textwidth]{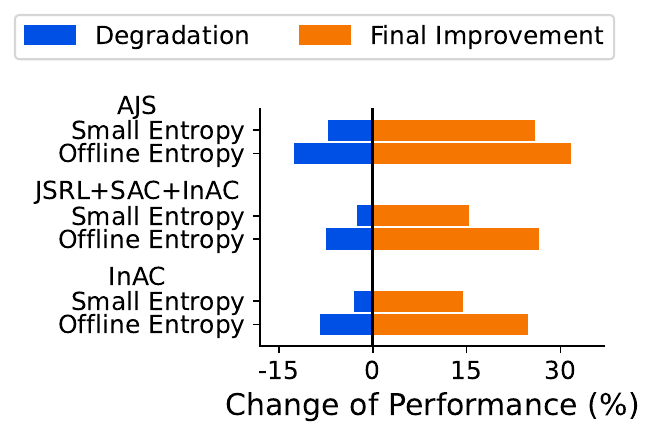}
    \caption{\label{fig:ajs_bar_small_ent} Performance differences when using a smaller entropy for InAC in fine-tuning. The experiment was conducted on HalfCheetah, Hopper, and Walker2d, using three different datasets from D4RL: Expert, Medium-Expert, and Medium. We tested 15 seeds for each of the 9 settings. 
    }
\end{wrapfigure}
AJS has two mechanisms for exploration: the entropy in the guide policy (InAC) and the exploration from SAC. The entropy in InAC is another source of hyperparameter tuning, and it worthwhile investigating how much AJS relies on this entropy. SAC uses auto-entropy tuning, but InAC uses the same fixed entropy used in offline training and in fine-tuning. Intuitively, the offline entropy can be higher, but once moving online, we want to avoid overly stochastic exploration and the potential for performance degradation. We test a setting here with a smaller entropy for InAC during fine-tuning.

In Figure \ref{fig:ajs_bar_small_ent} we can first see that InAC's final improvement drops significantly when we reduce the entropy, and its performance degradation also decreases slightly. This makes sense, as the policy explores less, and so learns more slowly with less degradation. AJS, on the other hand, has a much smaller decrease in the final improvement. This result does additionally highlight that allowing the guide policy to update with InAC, rather than freezing it to the offline policy as in the original JSRL algorithm, does provide algorithmic benefit. Restricting the exploration for the guide policy (InAC) resulted in slower learning. 
JSRL+SAC+InAC is like AJS, but without FQE to automatically reduce the exploration step $h$. We can see that it relies more on the exploration from InAC, than AJS. 



\subsection{Number of Exploration Steps}

We further investigated how the exploration step $h$ evolved during fine-tuning. 
The experiment was conducted using policies learned from the Expert and Medium datasets in Half Cheetah.

\begin{wrapfigure}[24]{t}{.5\textwidth}
    \centering
    \vspace{-0.3cm}
    \includegraphics[clip, trim={0 0 0 0}, width=0.5\textwidth]{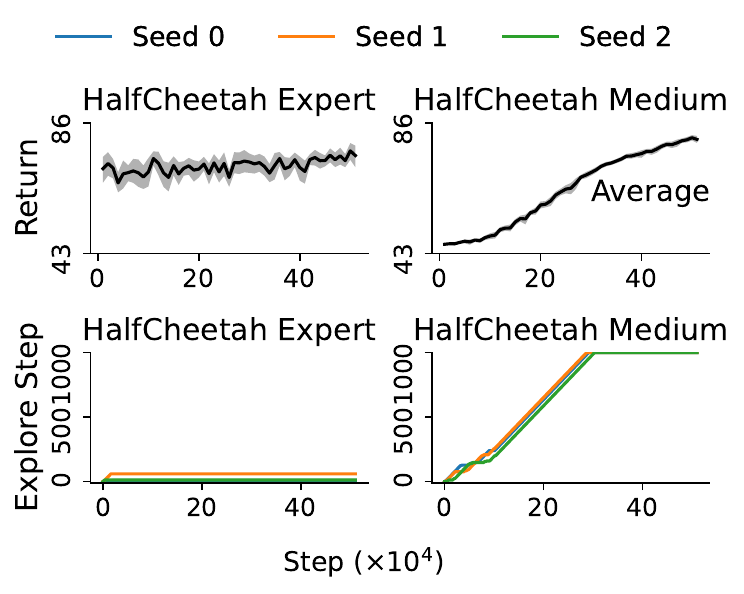}
    \vspace{-0.6cm}
    \captionof{figure}{With FQE, the exploration step increased more when the initial policy was worse. Fine-tuning was performed on Half Cheetah. We reported 3 seeds as examples of how $h$ was changed in a run. The two columns are for results using initial policies learned on Expert and Medium datasets separately. The y-axis in the first row indicates the return averaged on 15 seeds. The y-axis in the second row is the change in exploration steps in 500,000 fine-tuning steps. 
    The shaded area in the learning curve refers to the $95\%$ bootstrap confidence interval.
    }
    \label{fig:jsh_check}
\end{wrapfigure}
Figure \ref{fig:jsh_check} shows that FQE reduces the exploration steps as expected. 
For the Medium dataset in Figure \ref{fig:ajs_bar}, the offline policy is relatively poor. The exploration steps $h$ increased quickly, because the combined policy is better than this offline policy. The ASJ agent is able to quickly learn a better policy, rather than conservatively stepping back $h$ while being stuck running a poor guide policy. In contrast, for the Expert dataset where the offline policy is near-optimal, the agent explored conservatively, following the guide policy for longer. The exploration step stabilized at a relatively small value (less than 100) compared to the horizon of 1000 for this environment.

\section{Conclusion}

Balancing stability and policy improvement remains a challenge in fine-tuning. We demonstated that performance degradation exists in multiple offline-to-online algorithms, such as PEX, Proto and hybrid algorithms like using InAC offline and switching to SAC for fine-tuning. The degradation tends to be more severe when the offline policy is near-optimal, as apposed to when the agent starts from a suboptimal policy. More conservative algorithms, like using InAC in both offline training and fine-tuning, prevent performance degradation but learn too slowly afterwards. We proposed a new algorithm, called Automatic Jump Start (AJS), that leverages the stability of InAC and faster learning of SAC by slowly expanding the region controlled by SAC based on OPE estimates of performance. AJS was designed to avoid the need for hyperparameter tuning, towards the goal of practical offline-online algorithms for real-world applications where hyperparameter tuning is not possible.

\bibliography{main}
\bibliographystyle{rlj}





\beginSupplementaryMaterials
\section{Related Works}
The community has extensively explored how fine-tuning changes the neural network and its relationship to performance elevation. One observed pattern is that the top layers change more than the lower layers during fine-tuning \citep{peters-etal-2019-tune, merchant-etal-2020-happens, zhang2021revisiting, zhou-srikumar-2022-closer}. Additionally, fine-tuning alters the space of the network's hidden layers. \citet{zhou-srikumar-2022-closer} labels the representations using probing from a geometric perspective. The work points out that fine-tuning pushes the representations corresponding to different labels further apart. The pushed-away groups contribute to performance improvement during fine-tuning.

However, in reality, there remain issues with fine-tuning. Agents may suffer from severe performance degradation during the early fine-tuning stage. In literature, the issue has been observed and discussed in many works \citep{aghajanyan2021better, razdaibiedina2022improving, lyu2022mildly, lee2021offlinetoonline, song2023hybrid}. Various strategies have been proposed for stabilizing fine-tuning, such as collecting data with a relatively stable policy before updating the policy multiple times, instead of updating once per time step \citep{Julian2021never, Smith2022legged}, or making the offline data accessible during the online learning \citep{ball2023efficient}, but applying those methods solely does not fully prevent the performance degradation. The performance degradation results in additional time needed for the agent to improve the policy to match the initial performance level, thus reducing learning efficiency.

In the literature, various hypotheses and investigations were made to understand the performance degradation during fine-tuning. We classify them into three main categories:
1) \textbf{representation collapse and catastrophic forgetting} \citep{razdaibiedina2022improving, aghajanyan2021better,campos2021finetuning,zhang2023policy,song2023hybrid}: during fine-tuning, the network updates with respect to new samples, and fails to remember the policy or representation learned with offline data;
2) \textbf{distribution shift} \citep{lee2021offlinetoonline, zhao2022adaptive}: data collected from online interactions has a different distribution from the offline data, leading to severe bootstrap errors, thus distorting the learned function; and
3) \textbf{overestimation}: during offline learning, the action value can be overestimated \citep{lee2021offlinetoonline, nakamoto2023calql}. Agents suffer from bootstrapping errors when seeing online transitions and unlearn the pre-trained function.

The \emph{representation collapse} during fine-tuning has been observed and is believed to relate to performance degradation. According to the trust region theory, limiting changes in representation and preserving its generalizability mitigates the degradation \citep{aghajanyan2021better}.
Some works constrain the update on weight to ensure the policy does not change too rapidly, rather than directly constraining the representation change \citep{li2023proto, luo2023finetuning}. Similarly, \citet{razdaibiedina2022improving} addresses the problem with the multi-task learning setting. The work views the representation collapse as a form of overfitting to a single or a few tasks. The paper proposes to improve the representation's generalizability through pseudo-auxiliary tasks, which restrict changes in the representation structure instead of constraining weight updates.

\emph{Catastrophic forgetting} is considered another issue in fine-tuning and is said to be closely related to representation collapse \citep{aghajanyan2021better, razdaibiedina2022improving}. Approaches like Behavior Transfer and Policy Expansion aim to separate the behavior of the pre-trained policy and the newly learned policy \citep{campos2021finetuning, zhang2023policy}. In these methods, the pre-trained policy is fixed for exploitation and preventing catastrophic forgetting, while the newly learned policy focuses on exploration. Those papers examine the balance between adhering to the pre-trained policy and following exploratory actions. Moreover, \citet{song2023hybrid} suggests that giving the fine-tuning agent access to the offline data mitigates catastrophic forgetting.

The \emph{distribution shift} between the offline dataset and the data collected during fine-tuning has been discussed as another issue. Alleviating the sudden shift or preventing forgetting of the dataset distribution can improve the fine-tuning efficiency \citep{lee2021offlinetoonline, zhao2022adaptive, song2023hybrid, nair2021awac}. A straightforward approach is to incorporate offline data with online data, such as sampling the same amount of offline and online data in each batch \citep{ball2023efficient} or replacing uniform sampling with prioritized sampling: samples with higher online-ness are given higher priority \citep{lee2021offlinetoonline}. However, \citet{luo2023finetuning} presents a contrasting empirical finding in the fine-tuning of TD3-BC: TD3-BC enjoys a higher learning efficiency without initializing the buffer with offline data compared to feeding the offline data to the buffer.
Another solution to address the distribution shift is to consider existing off-policy learning algorithms. The distribution shift is also encountered in off-policy learning. Works such as \citet{ball2023efficient, luo2023finetuning} and \citet{nair2021awac} have been investigating how to transfer off-policy learning algorithms from the offline to the online setting.

Additionally, introducing constraints to policy updates has been found beneficial for stabilizing fine-tuning at the very beginning \citep{li2023proto}. In offline learning algorithms, various constraints are applied to the agent mainly to prevent bootstrapping from out-of-distribution actions and to mitigate overestimation. However, these issues are not typical in online learning, and the constraints slow down fine-tuning instead. While directly removing the constraint will cause performance degradation, the method proposed by \citet{zhao2022adaptive} learns adaptive weights for the constraint by monitoring the return over a short window and the current episode.

The overly \emph{conservatism} value estimation in offline learning can cause generalization issues and slow down online learning. The estimates can be arbitrarily lower than the actual value of a valid policy. At the beginning of fine-tuning, value estimations often need to increase to approach the true value \citep{nakamoto2023calql}. Several existing approaches attempt to address the issue. One method proposes to calibrate the value estimation of the learned policy to be higher than that of behavior policy \citep{nakamoto2023calql}. Another algorithm in \citet{lyu2022mildly} updates the out-of-distribution (OOD) action values toward a pseudo target, which is set to be a lower number than the maximum action value on the support set. As the pseudo target can be adjusted, the distance between the OOD action values and the maximum value on the support set is controlled, and the method can learn a mild conservative estimation.

A related issue is the conservatism introduced by the behavior model. Fitting the behavior model to the incoming data during fine-tuning is challenging \citep{Ramapuram2020lifelong}. \citet{nair2021awac} highlights that the inaccurate behavior model in fine-tuning causes conservative optimization. To address this, they propose AWAC, which implements an implicit policy constraint without relying on a behavior model.

Empirical evidence also supports that conservatism stabilizes fine-tuning. \citet{lee2021offlinetoonline} and \citet{nakamoto2023calql} used the conservative value estimation in offline learning to mitigate overestimation and obtained better performance in the online learning stage. Ensemble network offered a similar contribution. \citet{lee2021offlinetoonline} utilizes ensemble networks to enhance the conservatism of CQL pretraining. \citet{zhao2023improving} explicitly states that the ensemble mitigates performance degradation and presents stable fine-tuning performance with optimistic exploration. However, ensemble networks require large computational resources and suffer from slow updates regarding the wall clock time.

\section{Pseudocode}

This section includes the pseudocode of agents used for experiments.
For offline learning agents, the pseudocode of InAC is in Algorithm \ref{alg_inac}, the pseudocode of IQL is in Algorithm \ref{alg_iql}, and the pseudocode of SAC-based CQL is in Algorithm \ref{alg_cql}. The fine-tuning versions of InAC and IQL are in Algorithm \ref{alg_inac_ft} and Algorithm \ref{alg_iql_ft} separately. SAC's pseudocode is in Algorithm \ref{alg_sac}. PROTO's pseudocode is in Algorithm \ref{alg_proto}. Jump-Start's pseudocode is in Algorithm \ref{alg_js}. 
The pseudocode of AJS is in Algorithm \ref{alg_jsh}.

\begin{algorithm}[t]
    \caption{Automatic Jump-Start (AJS)}\label{alg_jsh}
    \begin{algorithmic}
        \STATE Initialize buffer $\mathcal{B}$ with offline data $\Dataset$
        \STATE Initialize guide policy $\pi_\eta$ using the offline learned policy
        \STATE Initialize exploration policy $\pi_\phi$ using the offline learned policy
        \STATE Initialize other networks (critic network, value network, etc) and the entropy $\tau$ with values obtained in offline training
        \STATE Define the Jump-Start policy $\pi_{js}$ following the function in Algorithm \ref{alg_jspolicy}, using $\pi_\eta$ as the guide policy and $\pi_\phi$ as the exploration policy
        \STATE Initialize max episode length or timeout $T$
        \STATE Initialize guide step $h=T$
        \STATE Initialize the reduction of guide step $\Delta= 2T / j$ where $j$ is the number of episode to run
        \STATE Select the initial states from the dataset and obtain the initial states set $S_0$
        \STATE Initialize the policy estimation function $F_\zeta$
        \STATE Initialize the number of iterations for initial OPE training $k$
        \STATE OPETraining($F_\zeta$, $\Dataset$, $\pi_{js}$, $k$)
        \STATE $v_{init} = F_\zeta(S_0, A_0)$

        \FOR{each episode}
        \STATE Initialize the step counter $t \leftarrow 1$
        \STATE Sample state $s$ from the environment

        \FOR{each step $t$}
        \STATE $a \sim \pi_{js}(s, t, h)$
        \STATE Interact with the environment to get $s', r$, and feed transition $(s, a, s', r)$ into the buffer $\mathcal{B}$

        \STATE Update $\pi_\phi$ and other networks used in SAC with SAC loss, including the entropy
        \STATE Update $\pi_\eta$ and other networks used in InAC with InAC loss

        \IF{There have been $T$ steps from the last update}
        \STATE OPETraining($F_\zeta$, $\mathcal{B}$, $\pi_{js}$, $T$)
        \ENDIF

        \ENDFOR

        \STATE $v_{ft} = F_\zeta(S_0, a \sim \pi_{js}(S_0, 0, h))$
        \IF{$v_{ft} \geq v_{init}$}
        \STATE $h \leftarrow \max(0, h - \Delta)$
        \ENDIF
        \STATE $s \leftarrow s'$; $t \leftarrow t + 1$

        \ENDFOR
    \end{algorithmic}
\end{algorithm}

\begin{wrapfigure}{R}{0.24\textwidth}
    \vspace{-0.5cm}
    \begin{minipage}{0.24\textwidth}
        \begin{algorithm}[H]
            \caption{JSPolicy}\label{alg_jspolicy}
            \begin{algorithmic}
                \STATE \textbf{Input} state $s$
                \STATE \textbf{Input} time step $t$
                \STATE \textbf{Input} guide step $h$
                \IF{$t > h$}
                \STATE $a \sim \pi_\phi(s)$
                \ELSE
                \STATE $a \sim \pi_\eta(\cdot|s)$
                \ENDIF
                \STATE \textbf{Return} $a$
            \end{algorithmic}
        \end{algorithm}
    \end{minipage}
\end{wrapfigure}

\begin{algorithm}
    \caption{OPETraining(FQE) (Algorithm 3 in \citet{le2019fqe})}
    \begin{algorithmic}
        \STATE \textbf{Input} function approximation $F_\zeta$
        \STATE \textbf{Input} dataset $\Dataset$
        \STATE \textbf{Input} evaluated policy $\pi_e$
        \STATE \textbf{Input} number of iterations $k$
        \STATE Maintain a target network $\bar{F}_\zeta$ and sync frequency $c$

        \FOR{i $\in$ [k]}
        \STATE Sample minibatch $\{s_i, a_i, r_i, s'_i\}$ from $\Dataset$
        \STATE Sample next action $a' \sim \pi_e(s')$
        \STATE Calculate the target $y = r + \gamma \bar{F}_\zeta(s', a')$
        \STATE Update $\zeta$ with MSE loss $\frac{1}{2} (F_\zeta(s, a) - y)^2$
        \STATE Sync target network every $c$ iterations
        \ENDFOR
    \end{algorithmic}
\end{algorithm}

\begin{algorithm}[t]
    \caption{InAC}\label{alg_inac}
    \textbf{Input:} Dataset $\Dataset$ with tuples of the form $(s, a, s', r)$; Weight $\tau$
    \begin{algorithmic}
        \STATE Initialize actor $\pi_\phi$
        \STATE Initialize critic $Q_\theta$
        \STATE Initialize value function $V_\psi$
        \STATE Initialize behavior policy simulator $\pi_\beta$
        \FOR{each gradient step}
        \STATE Update $\beta$ with loss $\mathbb{E}_{s,a \sim \Dataset} [-\log \pi_\beta(a|s)]$
        \STATE Update $\psi$ with loss $\mathbb{E}_{s \sim \Dataset, \hat{a}\sim \pi_\phi(\cdot|s)} [L_2(V_\psi(s) - (Q_\theta(s, \hat{a}) - \tau \log \pi_\phi(\hat{a}|s))) ]$
        \STATE Update $\theta$ with loss $\mathbb{E}_{s,a,s' \sim \Dataset, \hat{a'}\sim \pi_\phi(\cdot|s')} [L_2( Q_\theta(s,a) - ( r + \gamma (Q_{\bar{\theta}}(s', \hat{a'}) - \tau \log \pi_\phi(\hat{a'}|s')) ) ) ]$
        \STATE Update $\phi$ with loss $\mathbb{E}_{s,a \sim \Dataset} [-\exp{(\frac{Q_{\theta}(s, a) - V(s)}{\tau} - \pi_\beta(a|s))} \log \pi_\phi(a|s)$
        \ENDFOR
    \end{algorithmic}
\end{algorithm}

\begin{algorithm}[t]
    \caption{IQL}\label{alg_iql}
    \textbf{Input:} Dataset $\Dataset$ with tuples of the form $(s, a, s', r)$; Weight $\tau$
    \begin{algorithmic}
        \STATE Initialize actor $\pi_\phi$
        \STATE Initialize critic $Q_\theta$
        \STATE Initialize value function $V_\psi$
        \FOR{each gradient step}
        \STATE Update $\psi$ with loss $\mathbb{E}_{s \sim \Dataset, \hat{a}\sim \pi_\phi(\cdot|s)} [L_2^\rho (V_\psi(s) - Q_{\bar{\theta}}(s, \hat{a})) ]$, where $L_2^\rho(u)=|\rho - \mathbf{1}(u<0)| u^2$ and $\mathbf{1}$ represents an indicator function
        \STATE Update $\theta$ with loss $\mathbb{E}_{s,a,s' \sim \Dataset} [L_2 ( Q_\theta(s,a) - ( r + \gamma V_\psi(s') ) ) ]$
        \STATE Update $\phi$ with loss $\mathbb{E}_{s,a \sim \Dataset} [-\exp{(\frac{Q_{\bar{\theta}}(s, a) - V(s)}{\tau})} \log \pi_\phi(a|s) ]$
        \ENDFOR
    \end{algorithmic}
\end{algorithm}

\begin{algorithm}[t]
    \caption{CQL (SAC based)}\label{alg_cql}
    \textbf{Input:} Dataset $\Dataset$ with tuples of the form $(s, a, s', r)$
    \begin{algorithmic}
        \FOR{each time step}
        \STATE Calculate constaint $C = \alpha \mathbb{E}_{s\sim \Dataset} [\log \sum_a \exp(Q_\theta(s, a)) - \mathbb{E}_{a\sim \Dataset} [Q_\theta(s,a)]]$
        \STATE Update $\theta$ with loss $\mathbb{E}_{s,a,s' \sim \Dataset \hat{a'}\sim \pi_\phi(\cdot|s')} [L_2( Q_\theta(s,a) - ( r + \gamma (Q_{\bar{\theta}}(s', \hat{a'}) - \tau \log \pi_\phi(\hat{a'}|s')) ) ) ] + C$
        \STATE Update $\phi$ with loss $\mathbb{E}_{s \sim Dataset, \hat{a}\sim \pi_\phi(\cdot|s)} [\tau \log \pi_\phi(\hat{a}|s) - Q_\theta(s, \hat{a})]$
        \ENDFOR
    \end{algorithmic}
\end{algorithm}

\begin{algorithm}[t]
    \caption{InAC fine-tuning}\label{alg_inac_ft}
    \textbf{Input: } Weight $\tau$
    \begin{algorithmic}
        \STATE Initialize actor $\pi_\phi$ with offline learned actor
        \STATE Initialize critic $Q_\theta$ with offline learned critic
        \STATE Initialize value function $V_\psi$ with offline learned value function
        \STATE Initialize behavior policy simulator $\pi_\beta$ with offline learned behavior policy simulator
        \STATE Initialize buffer $\mathcal{B}$ with offline data $\Dataset$
        \FOR{each gradient step}
        \STATE Interact with the environment and feed transition $(s, a, s', r)$ into the buffer $\mathcal{B}$
        \STATE Update $\beta$ with loss $\mathbb{E}_{s,a \sim \Dataset} [-\log \pi_\beta(a|s)]$
        \STATE Update $\psi$ with loss $\mathbb{E}_{s \sim \Dataset, \hat{a}\sim \pi_\phi(\cdot|s)} [L_2(V_\psi(s) - (Q_\theta(s, \hat{a}) - \tau \log \pi_\phi(\hat{a}|s))) ]$
        \STATE Update $\theta$ with loss $\mathbb{E}_{s,a,s' \sim \Dataset, \hat{a'}\sim \pi_\phi(\cdot|s')} [L_2( Q_\theta(s,a) - ( r + \gamma (Q_{\bar{\theta}}(s', \hat{a'}) - \tau \log \pi_\phi(\hat{a'}|s')) ) ) ]$
        \STATE Update $\phi$ with loss $\mathbb{E}_{s,a \sim \Dataset} [-\exp{(\frac{Q_{\theta}(s, a) - V(s)}{\tau} - \pi_\beta(a|s))} \log \pi_\phi(a|s)$
        \ENDFOR
    \end{algorithmic}
\end{algorithm}

\begin{algorithm}[t]
    \caption{IQL fine-tuning}\label{alg_iql_ft}
    \textbf{Input:} Weight $\tau$
    \begin{algorithmic}
        \STATE Initialize actor $\pi_\phi$
        \STATE Initialize critic $Q_\theta$
        \STATE Initialize value function $V_\psi$
        \STATE Initialize buffer $\mathcal{B}$ with offline data $\Dataset$
        \FOR{each gradient step}
        \STATE Update $\psi$ with loss $\mathbb{E}_{s \sim \Dataset, \hat{a}\sim \pi_\phi(\cdot|s)} [L_2^\rho (V_\psi(s) - Q_{\bar{\theta}}(s, \hat{a})) ]$, where $L_2^\rho(u)=|\rho - \mathbf{1}(u<0)| u^2$ and $\mathbf{1}$ represents an indicator function
        \STATE Update $\theta$ with loss $\mathbb{E}_{s,a,s' \sim \Dataset} [L_2 ( Q_\theta(s,a) - ( r + \gamma V_\psi(s') ) ) ]$
        \STATE Update $\phi$ with loss $\mathbb{E}_{s,a \sim \Dataset} [-\exp{(\frac{Q_{\bar{\theta}}(s, a) - V(s)}{\tau})} \log \pi_\phi(a|s) ]$
        \ENDFOR
    \end{algorithmic}
\end{algorithm}

\begin{algorithm}[t]
    \caption{SAC}\label{alg_sac}
    \textbf{Input: } Weight $\tau$
    \begin{algorithmic}
        \STATE Initialize actor $\pi_\phi$ using the offline learned policy, \STATE Initialize critic $Q_\theta$ using the offline learned critic
        \STATE Initialize buffer $\mathcal{B}$ with offline data $\Dataset$
        \FOR{each time step}
        \STATE Interact with the environment and feed transition $(s, a, s', r)$ into the buffer $\mathcal{B}$
        \STATE Update $\theta$ with loss $\mathbb{E}_{s,a,s' \sim \mathcal{B}, \hat{a'}\sim \pi_\phi(\cdot|s')} [L_2( Q_\theta(s,a) - ( r + \gamma (Q_{\bar{\theta}}(s', \hat{a'}) - \tau \log \pi_\phi(\hat{a'}|s')) ) ) ]$
        \STATE Update $\phi$ with loss $\mathbb{E}_{s \sim \mathcal{B}, \hat{a}\sim \pi_\phi(\cdot|s)} [\tau \log \pi_\phi(\hat{a}|s) - Q_\theta(s, \hat{a})]$
        \ENDFOR
    \end{algorithmic}
\end{algorithm}

\begin{algorithm}[t]
    \caption{PROTO (SAC based)}\label{alg_proto}
    \textbf{Input: } Weights $\tau$ and $\alpha$
    \begin{algorithmic}
        \STATE Initialize actor $\pi_\phi$ using the offline learned policy
        \STATE Initialize critic $Q_\theta$ using the offline learned critic
        \STATE Initialize buffer $\mathcal{B}$ with offline data $\Dataset$
        \FOR{each time step}
        \STATE Interact with the environment and feed transition $(s, a, s', r)$ into the buffer $\mathcal{B}$
        \STATE Update $\theta$ with loss $\mathbb{E}_{s,a,s' \sim \mathcal{B}, \hat{a'}\sim \pi_\phi(\cdot|s')} [L_2( Q_\theta(s,a) - ( r + \gamma (Q_{\bar{\theta}}(s', \hat{a'}) - \tau \log \pi_\phi(\hat{a'}|s')) - \alpha \log \frac{\pi_\phi(\hat{a'}|s')}{\pi_{\bar{\phi}}(\hat{a'}|s')} ) ) ]$
        \STATE Update $\phi$ with loss $\mathbb{E}_{s \sim \mathcal{B}, \hat{a}\sim \pi_\phi(\cdot|s)} [\tau \log \pi_\phi(\hat{a}|s) - Q_\theta(s, \hat{a})  + \alpha \log \frac{\pi_\phi(\hat{a}|s)}{\pi_{\bar{\phi}}(\hat{a}|s)}]$
        \ENDFOR
    \end{algorithmic}
\end{algorithm}

\begin{algorithm}[t]
    \caption{Jump-Start (SAC based)}\label{alg_js}
    \begin{algorithmic}
        \STATE Initialize buffer $\mathcal{B}$ with offline data $\Dataset$
        \STATE Initialize guide policy $\pi^g$ using the offline learned policy
        \STATE Initialize exploration policy $\pi_\phi$ using the offline learned policy
        \STATE Initialize critic $Q_\theta$ using the offline learned critic
        \STATE Initialize guide-step $h \leftarrow T$ where $T$ is the number of steps of one episode.
        \STATE Set the reduction value of guide-step $p$
        \STATE Set performance threshold $\epsilon$

        \FOR{each episode}
        \STATE Sample state $s$ from the environment; Initialize counter $t\leftarrow 1$
        \FOR{each step in episode}
        \IF{$t>h$}
        \STATE $a \sim \pi^e(\cdot|s)$
        \ELSE
        \STATE $a \sim \pi_\phi(\cdot|s)$
        \ENDIF
        \STATE Interact with the environment and feed transition $(s, a, s', r)$ into the buffer $\mathcal{B}$
        \STATE $s \leftarrow s'$ ; $t \leftarrow t+1$
        \ENDFOR
        \FOR{each step in episode}
        \STATE Update $\theta$ with loss $\mathbb{E}_{s,a,s' \sim \mathcal{B}, \hat{a'}\sim \pi_\phi(\cdot|s')} [L_2( Q_\theta(s,a) - ( r + \gamma (Q_{\bar{\theta}}(s', \hat{a'}) - \tau \log \pi_\phi(\hat{a'}|s')) ) ) ]$
        \STATE Update $\phi$ with loss $\mathbb{E}_{s \sim \mathcal{B}, \hat{a}\sim \pi_\phi(\cdot|s)} [\tau \log \pi_\phi(\hat{a}|s) - Q_\theta(s, \hat{a})]$
        \ENDFOR
        \IF {The performance of the latest episode is better than $1-\epsilon$ of the previous best performance}
        \STATE $h \leftarrow max(0, h - p)$
        \ENDIF
        \ENDFOR
    \end{algorithmic}
\end{algorithm}

\begin{algorithm}[t]
    \caption{Jump-Start fixed schedule (SAC based)}\label{alg_js_fix_schedule}
    \begin{algorithmic}
        \STATE Initialize buffer $\mathcal{B}$ with offline data $\Dataset$
        \STATE Initialize guide policy $\pi^g$ using the offline learned policy
        \STATE Initialize exploration policy $\pi_\phi$ using the offline learned policy
        \STATE Initialize critic $Q_\theta$ using the offline learned critic
        \STATE Initialize guide-step $h \leftarrow T$ where $T$ is the number of steps of one episode.
        \STATE Set the reduction value of guide-step $p$
        \FOR{each episode}
        \STATE Sample state $s$ from the environment; Initialize counter $t\leftarrow 1$
        \FOR{each step in episode}
        \IF{$t>h$}
        \STATE $a \sim \pi^e(\cdot|s)$
        \ELSE
        \STATE $a \sim \pi_\phi(\cdot|s)$
        \ENDIF
        \STATE Interact with the environment and feed transition $(s, a, s', r)$ into the buffer $\mathcal{B}$
        \STATE $s \leftarrow s'$ ; $t \leftarrow t+1$
        \STATE Update $\theta$ with loss $\mathbb{E}_{s,a,s' \sim \mathcal{B}, \hat{a'}\sim \pi_\phi(\cdot|s')} [L_2( Q_\theta(s,a) - ( r + \gamma (Q_{\bar{\theta}}(s', \hat{a'}) - \tau \log \pi_\phi(\hat{a'}|s')) ) ) ]$
        \STATE Update $\phi$ with loss $\mathbb{E}_{s \sim \mathcal{B}, \hat{a}\sim \pi_\phi(\cdot|s)} [\tau \log \pi_\phi(\hat{a}|s) - Q_\theta(s, \hat{a})]$
        \ENDFOR
        \STATE $h \leftarrow max(0, h - p)$
        \ENDFOR
    \end{algorithmic}
\end{algorithm}

\section{Ensemble Network Mitigates Overestimation but is Hard to Tune} \label{apdx:ensemble}

\begin{figure}[th]
    \centering
    \subfloat[][Learn from the minimum estimation\label{fig:ensemble_min}]{\includegraphics[width=0.75\textwidth]{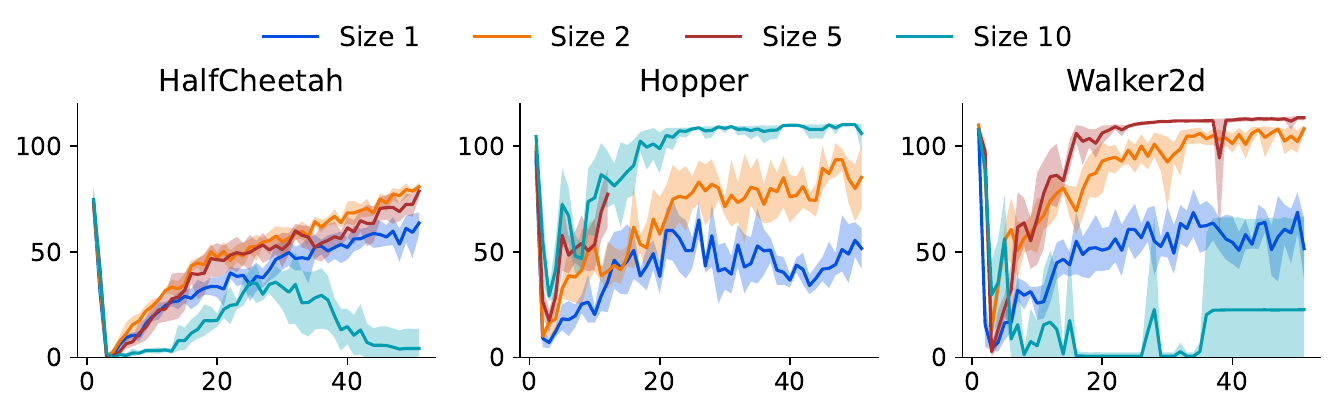}}

    \subfloat[][Learn from the median estimation\label{fig:ensemble_median}]{\includegraphics[width=0.75\textwidth]{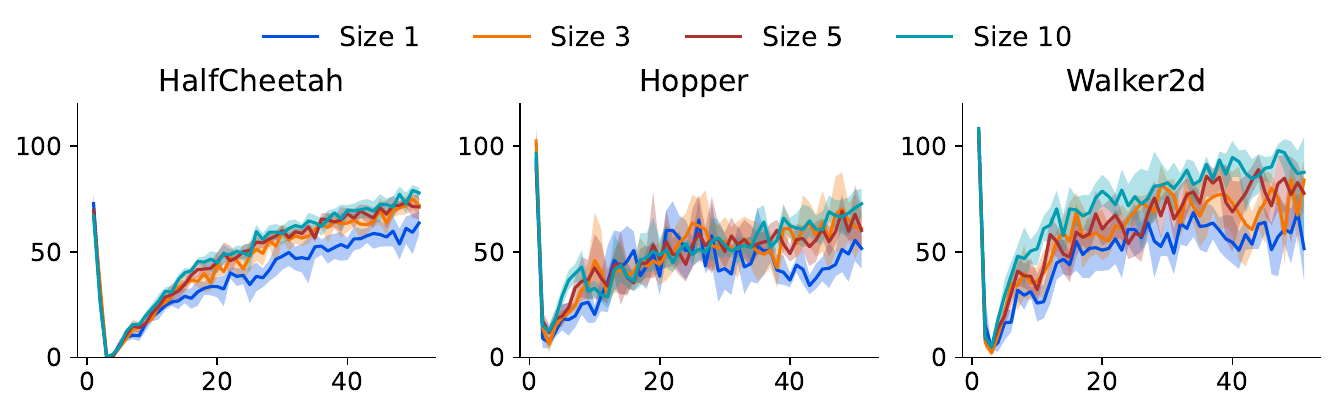}}
    \vspace{-0.3cm}
    \caption{\label{fig:ensemble}{
            Taking the minimum estimation in ensemble networks for bootstrapping (the first subplot), a larger ensemble size has a worse fine-tuning performance in 2 out of 3 cases. In Hopper, a larger ensemble size learns faster and has an improved worst performance, compared to sizes 1 and 2, but the performance degradation still exists. When taking the median estimation in ensemble networks (the second subplot), size 10 learns faster than the smaller size, though there remained no improvement in the performance degradation.
            The three columns present the performance in HalfCheetah, Hopper, and Walker2D separately. The x-axis is the time step $(\times 10^4)$, and the y-axis is the normalized performance. The blue, orange, green, and red curves indicate the performance of using 1, 2, 5, and 10 networks in the first row, and 1, 3, 5, and 10 networks in the second row. The shaded area refers to $95\%$ bootstrap confidence interval.}
    }
\end{figure}

\begin{figure}[th]
    \centering
    \includegraphics[width=0.75\textwidth]{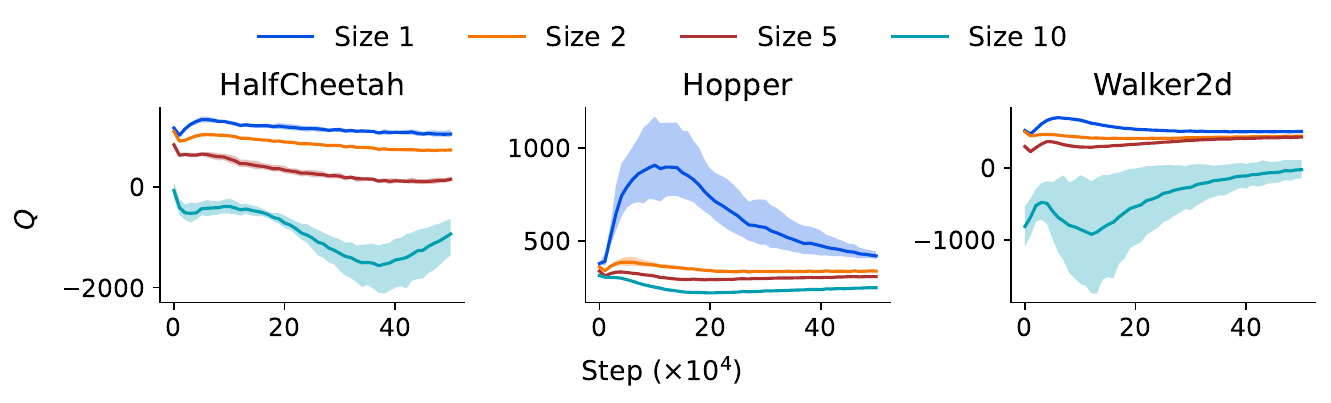}
    \includegraphics[width=0.75\textwidth]{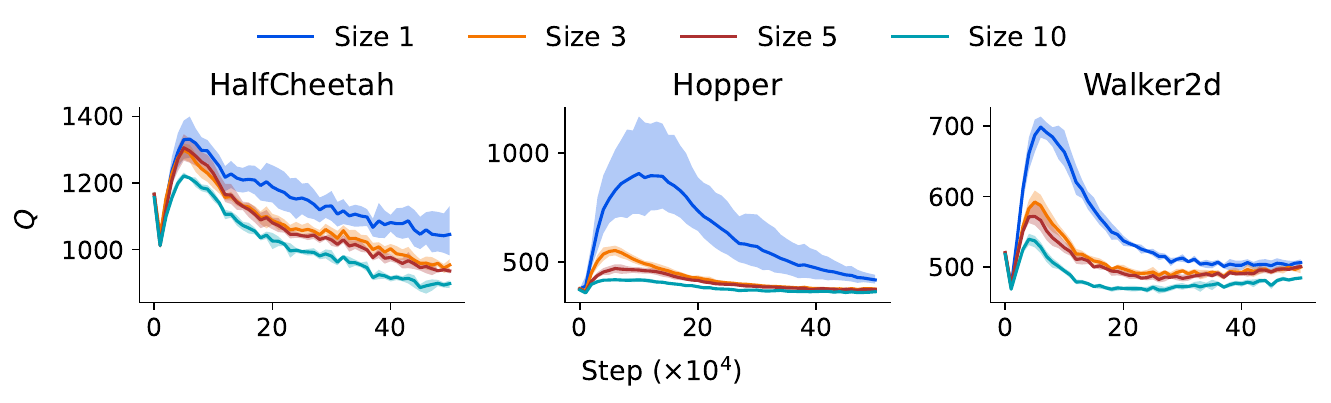}
    \vspace{-0.3cm}
    \caption{\label{fig:ensemble_q}{When the agent bootstraps from the minimum estimate of the ensemble network (the first row), the learned action value decreases as the ensemble size increases. In the size 10 setting, the minimum action value estimate is overly conservative. When the agent bootstraps from the median estimate of the ensemble network, the learned action value decreases as the ensemble size increases. There was no over-pessimistic estimate in the tested sizes.
            Each subplot provides the action value estimates by various ensemble sizes in HalfCheetah, Hopper, and Walker2D separately. The blue, orange, and green curves are the minimum estimates using 1, 2, 5, and 10 networks in the first row. They are the median estimates using 1, 3, 5, and 10 networks in the second row. The x-axis is the number of time steps $(\times 10^4)$, and the y-axis is the learned action value.}
    }
\end{figure}

\begin{figure}[th]
    \centering
    \includegraphics[width=0.8\textwidth]{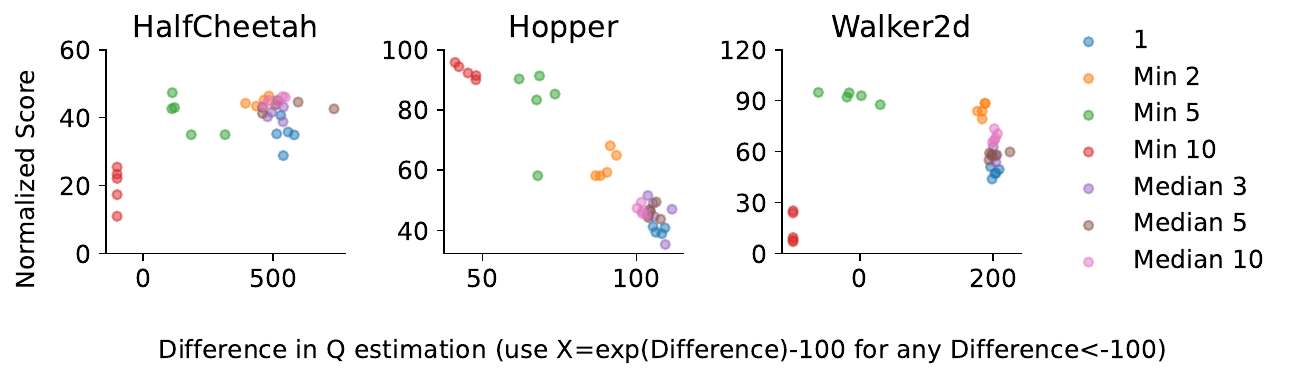}
    \includegraphics[width=0.8\textwidth]{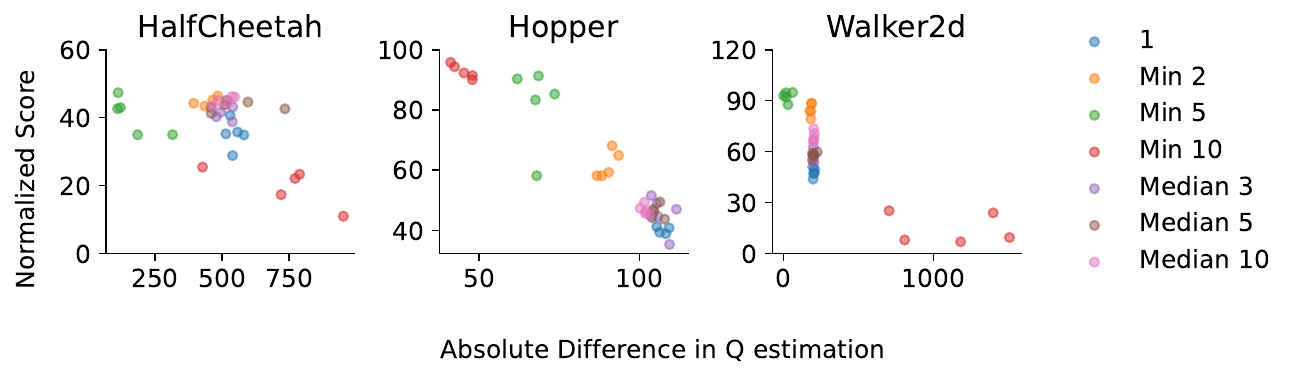}
    \vspace{-0.3cm}
    \caption{\label{fig:ensemble_q_stat}{Both over-optimistic and over-pessimistic estimates hurts the performance. In both subplots, the y-axis is the normalized score. The x-axis of the first subplot is the difference between the action value estimate and the true discounted return in rollout. For a better visualization, the x-coordinate of scatters is projected to $\exp(x)-100$ if the difference is smaller than -100. In the second subplot, the x-axis is the absolute difference.}
    }
\end{figure}

Earlier studies have suggested that learning with ensemble networks enhances conservatism and contributes to stabilizing fine-tuning \cite{lee2021offlinetoonline,zhao2023improving}. A sufficiently large ensemble network alleviates the overestimation~\citep{an2021uncertainty,zhao2023improving}. In this section, we investigate how ensemble critic networks affect the fine-tuning performance of an InAC policy and point out that the ensemble strategy does not always work without careful tuning. Experiments in this section took the best entropy in $\{0.33, 0.1, \textit{AUTO}\}$. We swept the parameter using 5 seeds, then reported an extra 15 seeds for the best setting only.

The empirical results highlighted the possibility of using an ensemble network to improve learning efficiency by gaining a conservative action value estimation. The improvement, however, comes with the condition that the value estimation should not be overly pessimistic. We found that a large ensemble size may suffer from overly pessimistic value estimation. In this case, the ensemble critic slows down fine-tuning, instead of improving the performance. We further measured the degree of conservatism in the value estimation, and concluded that staying close to the true value should be required when pursuing conservative value estimation. The experimental results demonstrated that a larger ensemble size did not improve performance degradation.

We tested two different bootstrapping methods. In the first test, the agent bootstrapped from the minimum estimate in ensemble networks. Figure \ref{fig:ensemble} illustrates that increasing the ensemble size cannot consistently improve the fine-tuning performance. In HalfCheetah and Walker2d, the performance of size-10-ensemble quickly dropped and failed to recover back to the initial level in $700,000$ steps. A smaller size (2) learned faster and converged to better policy, even though there was a severe performance degradation at the beginning. Hopper had a different pattern: an increasing ensemble size empirically showed an improving performance.

The second test required the agent to bootstrap from the median estimate in ensemble critic. Increasing the number of networks to 10 did not hurt the fine-tuning performance as in the first test. Instead, the experiment result in Figure \ref{fig:ensemble} suggested the learning efficiency increased monotonically when the ensemble size increased from 1 to 10. When checking the change of value estimates, we noticed that a larger ensemble network also had a lower estimate, as when using the minimum for bootstrapping, but we did not observe any estimate staying lower than zero.

In further investigation, we conclude that the inconsistent performance is related to the accuracy of the action value estimation. We visualize the minimum value estimate in the ensemble critic in Figure \ref{fig:ensemble_q}. In the two environments, HalfCheetah and Walker2d, where a size-10 ensemble experienced failure, we observed an overly conservative value estimate. After the fine-tuning started, the value estimates of sizes 1 and 2 experienced a smaller change than size 10. In comparison, the value estimate of size 10 had a larger change than the other two settings and converged more slowly. In Hopper, the initialized value estimate of all 3 settings remained in a similar range, while the estimate of a larger ensemble size turned out to be smaller. The patterns of estimate's change after fine-tuning starts of all 3 settings were consistent, while size 10 maintained a lower estimate than the other two settings.

The estimation learned with median value bootstrapping stayed in a reasonable range (Figure \ref{fig:ensemble_q}). The estimates monotonically decreased as the ensemble size decreased. The difference across ensemble sizes turned out to be smaller than the difference in minimum value bootstrapping.

Therefore, a more conservative estimation cannot imply better fine-tuning. We attempt to figure out the threshold of gaining improvement using an ensemble. Figure \ref{fig:ensemble_q_stat} suggested the correlation between the error in action value estimates and the fine-tuning performance. We checked the difference between the learned value estimation $\hat{Q}$ and the true discounted return by deploying the offline learned policy $G$. The true return was estimated by averaging the return obtained in 5 rollouts with 1000 steps each. The discount value in rollout remained the same during offline training (0.99). We noticed that as $\hat{Q} - G$ decreased, the performance first increased, then decreased. When checking the absolute value, $|\hat{Q} - G|$, we noticed an increasing fine-tuning performance as the absolute difference decreased.

Our results further highlighted the importance of maintaining accurate value estimation in a reasonable range even when pursuing pessimism. If the estimation is overly pessimistic, the fine-tuning performance will be hurt. However, controlling the range of Q estimates is not straightforward when applying ensemble architecture in offline learning. One obvious reason is that the threshold, the true return, is usually unknown, and the accessibility to the true environment is limited.

\section{Testing for Linear Paths Between the Offline Policy to a Better Policy}\label{sec:linear_comb}
Fully preventing performance degradation is non-trivial because of the limited knowledge the agent has on the space of policy and performance. An offline learned policy could be a local optima. Moving out of the local optima means the policy may pass through an area where the performance gets worse before it finds a better optima. Following the shortest path between the policy initialization and the final policy does not fully prevent the degradation.

\begin{figure}[th]
    \centering
    \captionsetup[subfloat]{margin=3pt,format=hang}

    \subfloat[][Expert]{\includegraphics[width=\textwidth]{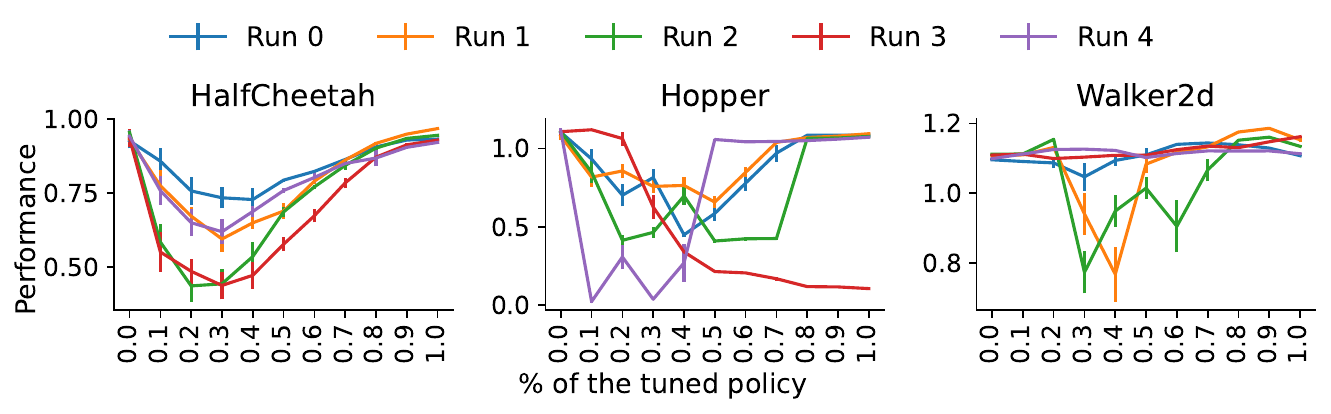}\label{fig:linear_combine_expert}}
    
    \subfloat[][Medium-Expert]{\includegraphics[width=\textwidth]{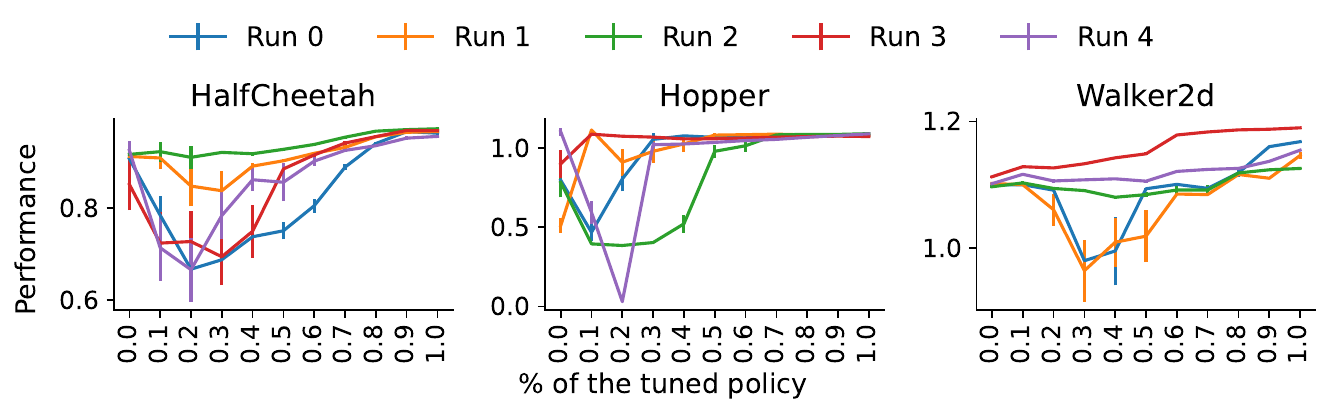}\label{fig:linear_combine_medexp}}
    
    \subfloat[][Medium]{\includegraphics[width=\textwidth]{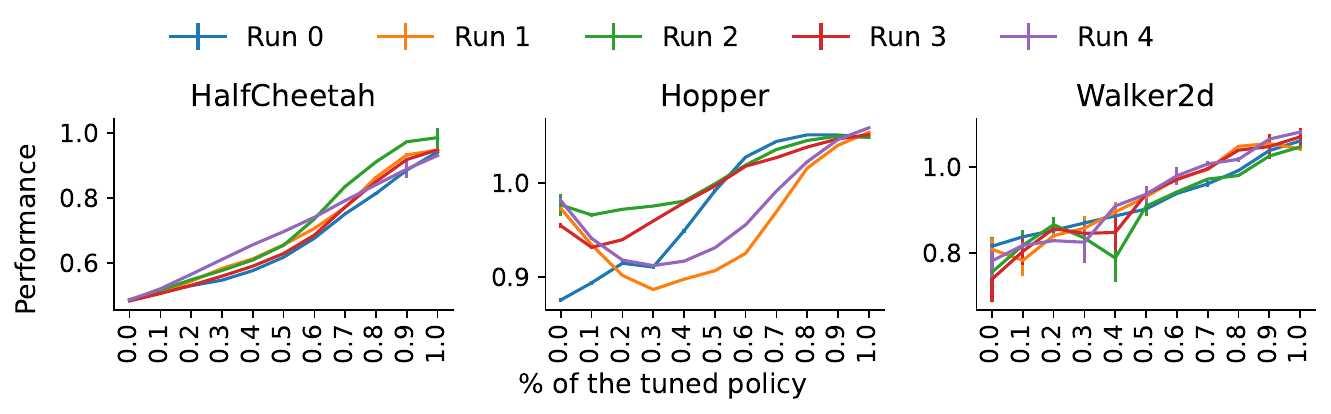}\label{fig:linear_combine_medium}}
    \caption{{With a policy initialization derived from the Expert dataset, linearly updating the initialization to SAC's final policy did not ensure a monotonic increasing performance. When learning from the Medium-Expert dataset, curves had less degradation than using the Expert dataset. When learning with a Medium dataset, curves had smaller degradation than the other two cases. Each evaluation was done with 50 rollout trajectories with a timeout after 1000 steps. The three columns show the result in HalfCheetah, Hopper, and Walker2D separately. In each subplot, each curve shows the performance of one random seed. The x-axis is the ratio of SAC's final policy, and the y-axis is the normalized return in evaluation. The $95\%$ confidence interval is indicated with the vertical error bar. The leftmost performance in each subplot is the normalized return of the policy initialization, and the rightmost is the normalized return of SAC's final policy after fine-tuning. Moving from left to right, the ratio of SAC's final policy increases. }
    }
\end{figure}

To visualize how the performance changes on the shortest path between the starting and the final policies at fine-tuning, we simulated the update with a linear combination of the policy initialization and the final policy learned by SAC. The focus was kept on using the near-optimal dataset to learn a policy initialization. We examined how the performance changed with a linear combination of the two policies, with a changing ratio. Figure \ref{fig:linear_combine_expert} indicated that when linearly combining the two policies with a $70\%$ and $30\%$ ratio, the performance degraded to a near-random level (below $0.3$), even though the policy initialization was always near-optimal. The performance got close to the optimal level only after the ratio of SAC's final policy went above around $80\%$.

When using the Medium dataset for offline policy learning, the policy initialization had a smaller degradation compared to the policy learned from the near-optimal dataset. In Figure \ref{fig:linear_combine_medium}, we observed an improved worst performance during the linear shifting, which was always above $0.3$, even though the policy initialization (the leftmost point) had a worse performance.


\section{Additional Results} \label{apdx:results}

\subsection{Different Entropy Settings}
\begin{figure}[th]
    \centering
    \includegraphics[width=0.85\textwidth]{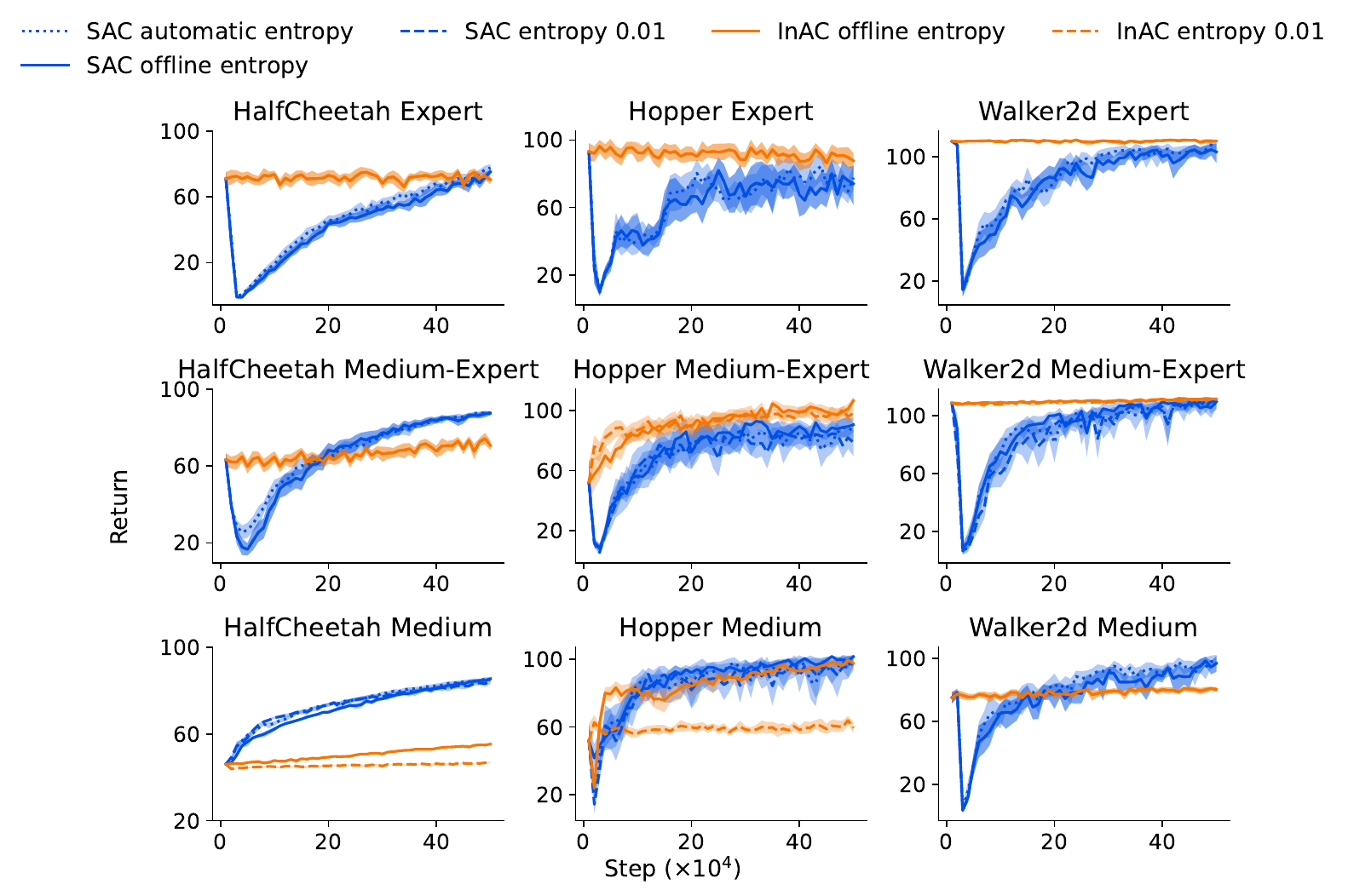}
    \caption{\label{fig:ft_inac_small_ent}{Fixing the entropy to a small value ($0.01$) restricted the policy improvement in InAC, while SAC was not influenced much. The x-axis refers to the time steps. The y-axis is the normalized return. The shaded area is the $95\%$ bootstrap confidence interval.}}
\end{figure}

Figure \ref{fig:ft_inac_small_ent} demonstrates the following results: (1) we confirmed that setting a fixed small constant (0.01) as InAC's entropy was good enough to maintain the stability; and (2) using a fixed entropy for SAC, no matter the offline setting or a small constant (0.01), did not mitigate performance degradation and was no better than the automatic entropy tuning SAC.

\subsection{Offline Learning Performance}
\begin{figure}[th]
    \centering
    \subfloat[][Ensemble size 1]{\includegraphics[width=0.75\textwidth]{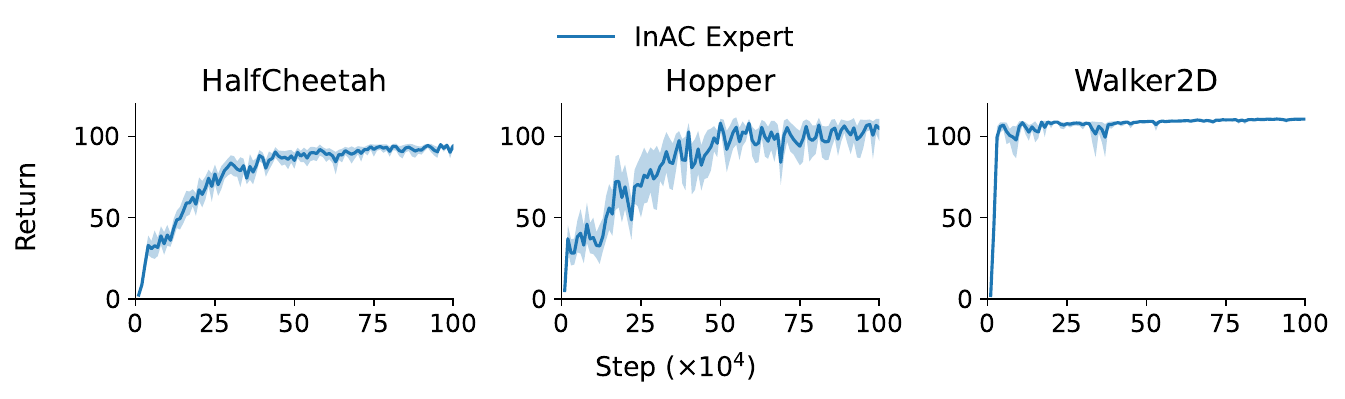}};
    \subfloat[][Ensemble size 2]{\includegraphics[width=0.75\textwidth]{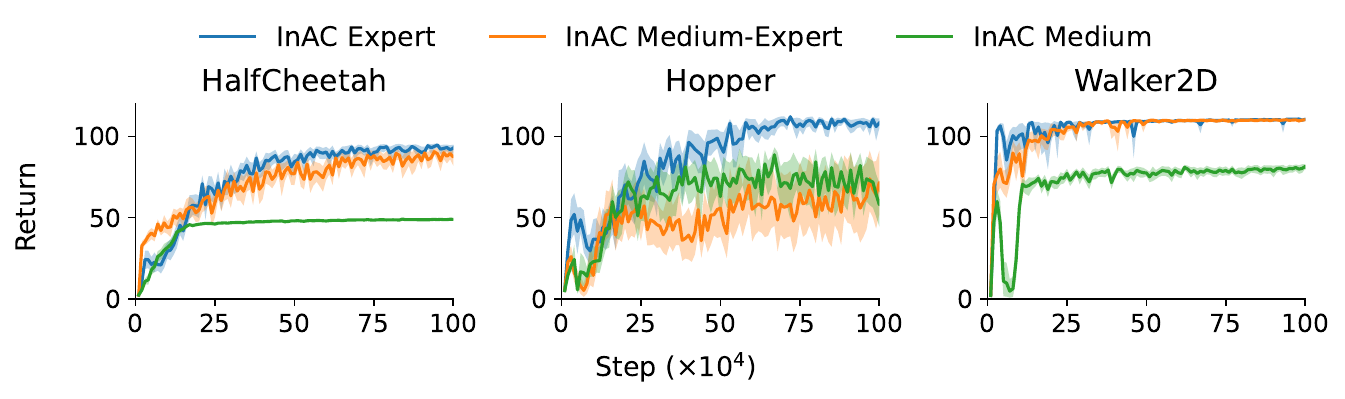}};
    \subfloat[][Ensemble size 10]{\includegraphics[width=0.75\textwidth]{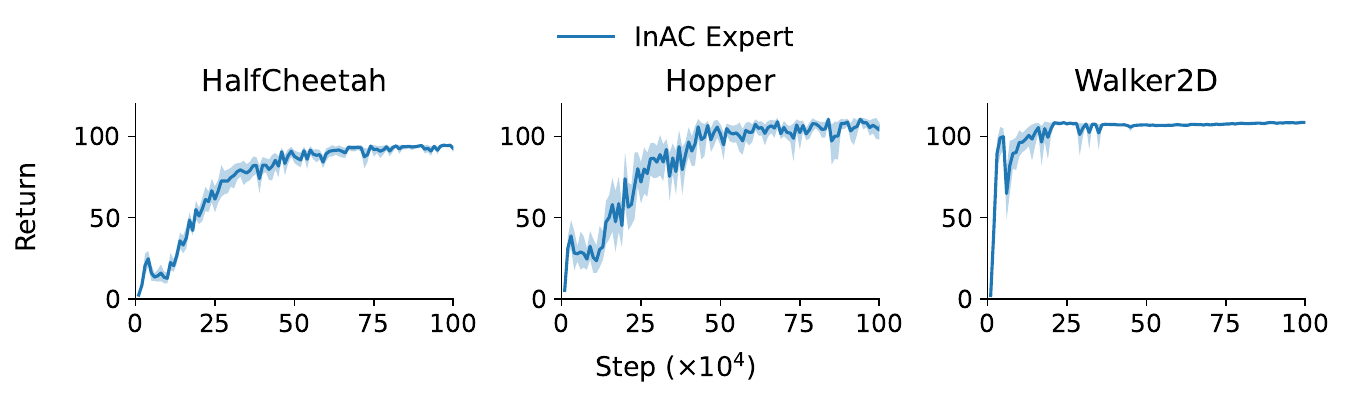}}
    \caption{\label{fig:apdx_offline_inac}{InAC converged in offline learning. The x-axis is the number of updates. The y-axis is the normalized return. The shaded area refers to $95\%$ bootstrap confidence interval.}}
\end{figure}

\begin{figure}[th]
    \centering
    \includegraphics[width=0.55\textwidth]{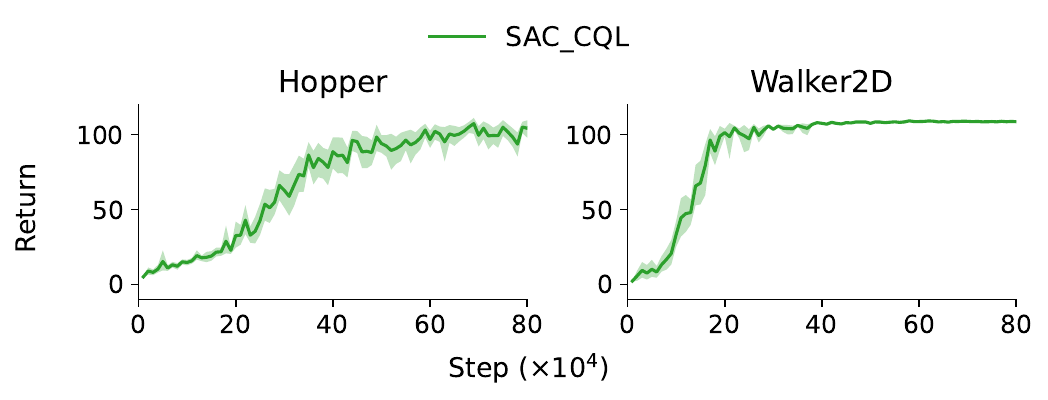}
    \caption{\label{fig:apdx_offline_sac_cql}{SAC-based CQL with 10 critics converged in offline learning. The x-axis is the number of updates. The y-axis is the normalized return. The shaded area refers to $95\%$ bootstrap confidence interval.}}
\end{figure}

\begin{figure}[th]
    \centering
    \includegraphics[width=0.75\textwidth]{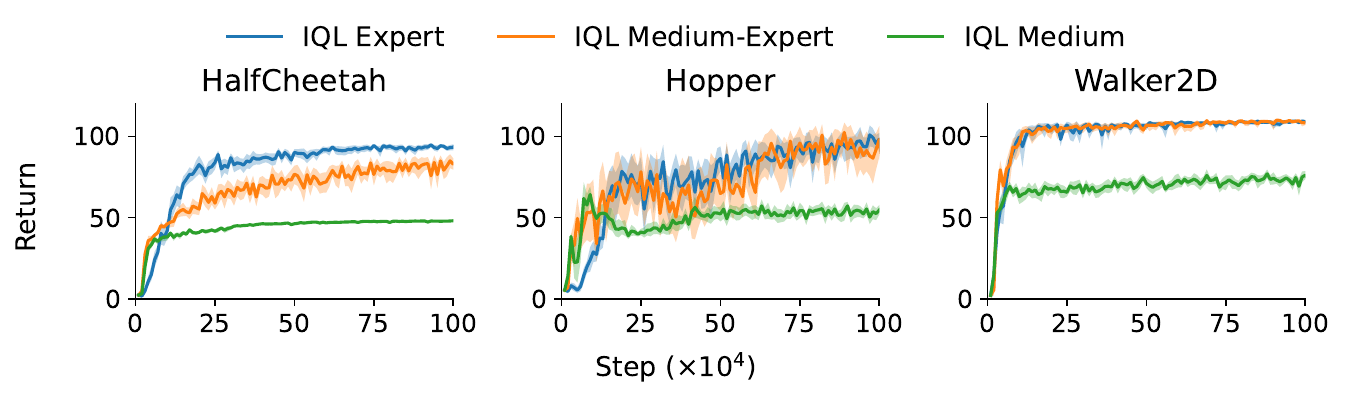}
    \caption{\label{fig:apdx_offline_iql}{IQL converged in offline learning. The x-axis is the number of updates. The y-axis is the normalized return. The shaded area refers to $95\%$ bootstrap confidence interval.}}
\end{figure}

We put learning curves of InAC with an ensemble size of 2 in Figure \ref{fig:apdx_offline_inac}. Learning curves of SAC+CQL and SAC, both with 10 critics, are in Figure \ref{fig:apdx_offline_sac_cql}.
Learning curves of IQL with an ensemble size of 2 are in Figure \ref{fig:apdx_offline_iql}.

\subsection{Fixed Schedule Jump-Start} \label{apdx:fix_schedule_js}

\begin{figure}[th]
    \centering
    \includegraphics[clip, trim={0 0 8cm 0}, width=0.75\textwidth]{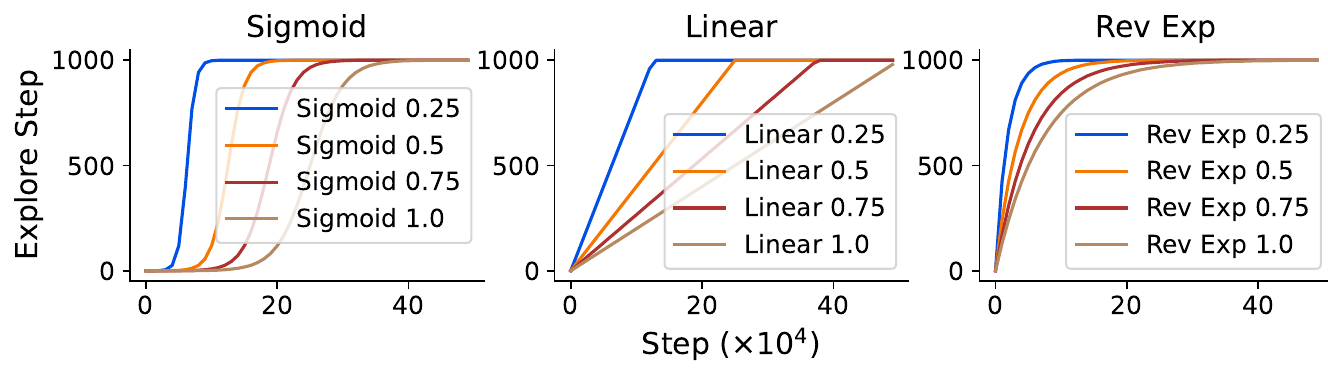}
    \caption{\label{fig:fix_schedule_curve}{The figure indicates how the number of exploration steps changed with different schedules and parameters. The y-axis is the number of explore steps, and the x-axis is the time steps.}}
\end{figure}

\begin{figure}[th]
    \centering
    \includegraphics[width=0.75\textwidth]{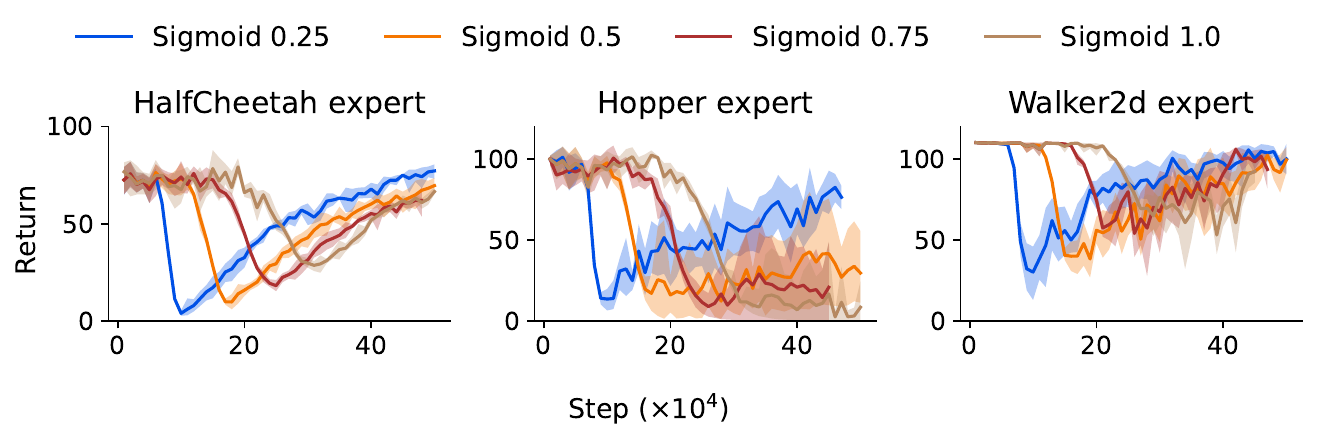}
    \includegraphics[width=0.75\textwidth]{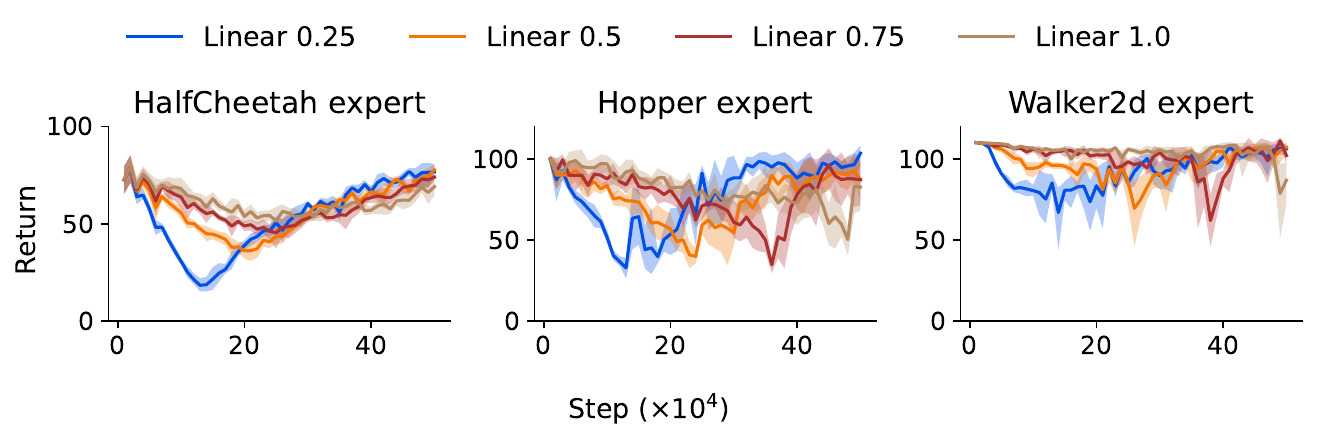}
    \includegraphics[width=0.75\textwidth]{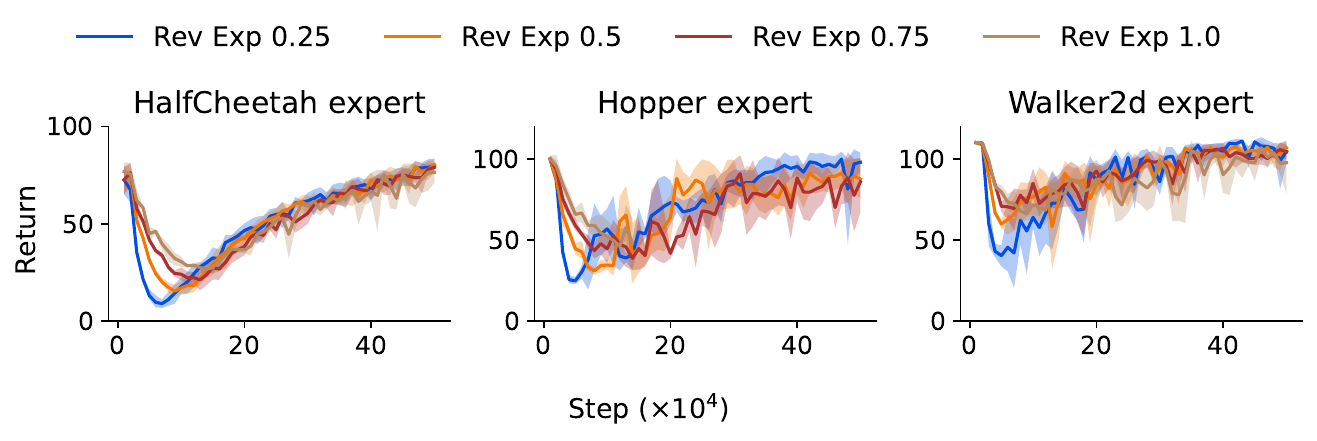}
    \caption{\label{fig:fix_schedule_lc}{Performance degradation existed if using a fixed schedule to increase the explore steps. The y-axis is the normalized return. The x-axis is the time steps. The shaded area refers to $95\%$ bootstrap confidence interval.}}
\end{figure}

It remains hard to search for a fixed exploration steps schedule to prevent performance degradation. We tested several fixed schedules as listed below:
\begin{enumerate}
    \item Sigmoid: The number of exploration steps increases following a sigmoid curve. The speed of expanding exploration remains slow at the beginning, increasing in the middle of the run, then decreases.
    \item Linear: The number of exploration steps increases linearly.
    \item Rev Exp: The number of exploration steps increases following a reversed exponential curve. The exploration step expands fast at the beginning, and the speed of expansion decreases later.
\end{enumerate}
The shape of each curve is controlled by an extra parameter. A larger value means the explore step increases slower. We tested $\{0.25, 0.5, 0.75, 1.0\}$.
The learning curves are shown in Figure \ref{fig:fix_schedule_lc}. We put the corresponding curves for the change of exploration steps in Figure \ref{fig:fix_schedule_curve}. All settings suffered from severe performance degradation.

\subsection{AJS Learning Curves}
\begin{figure}[th]
    \centering
    \includegraphics[width=1\textwidth]{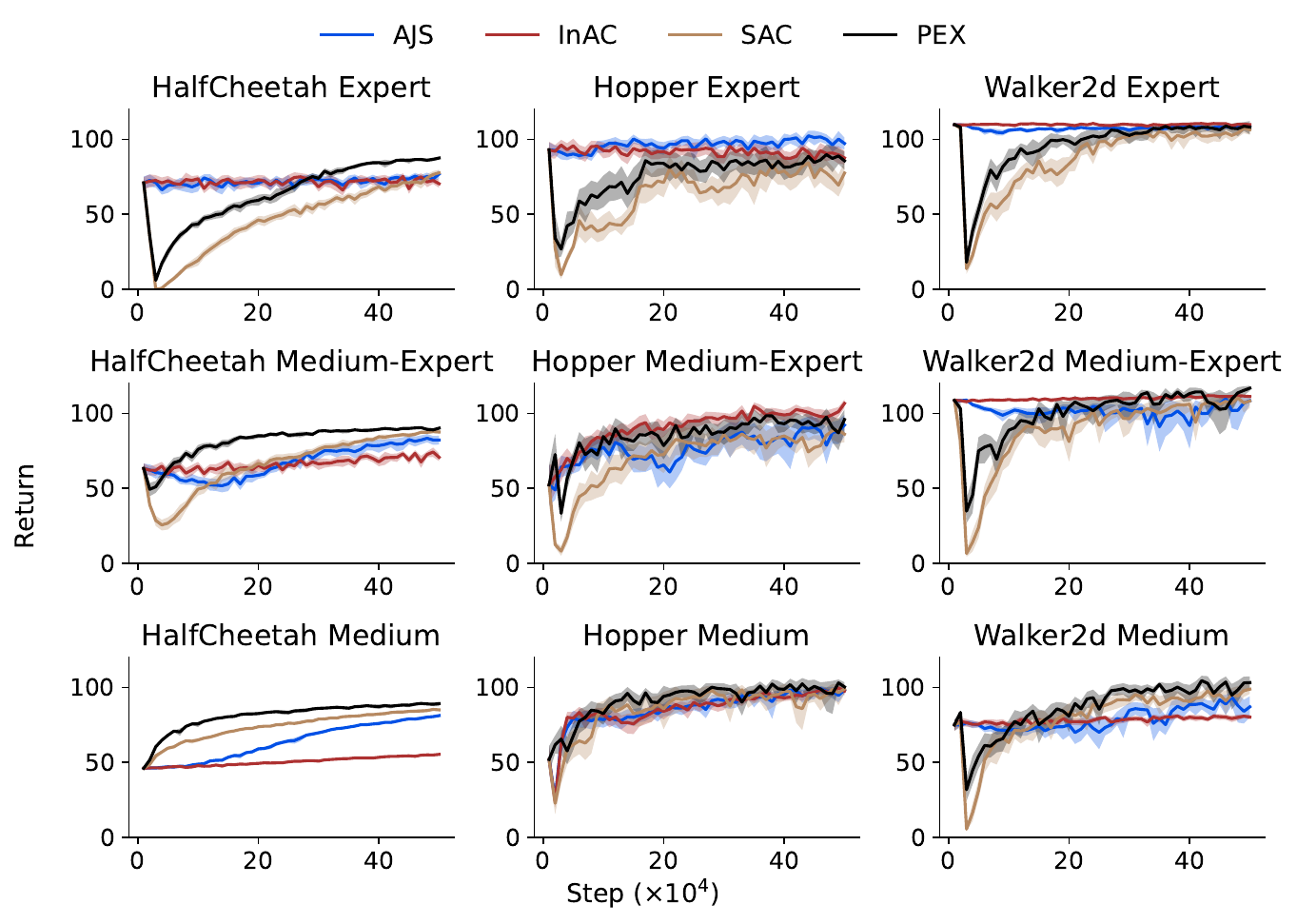}
    \vspace{-0.3cm}
    \caption{\label{fig:ajs_curve}{AJS mitigated performance degradation while maintaining the ability to explore policy improvement. The x-axis is the time step. The y-axis is the normalized return. Different columns include the learning curves in HalfCheetach, Hopper, and Walker2D, respectively. The first row contains the result given the policy initialization learned from the Expert dataset. The second and third rows are initializations learned from Medium-Expert and Medium datasets, respectively. Performance is averaged on 15 seeds. The shaded area refers to $95\%$ bootstrap confidence interval.
        }}
\end{figure}

\begin{figure}[th]
    \centering
    \includegraphics[width=1\textwidth]{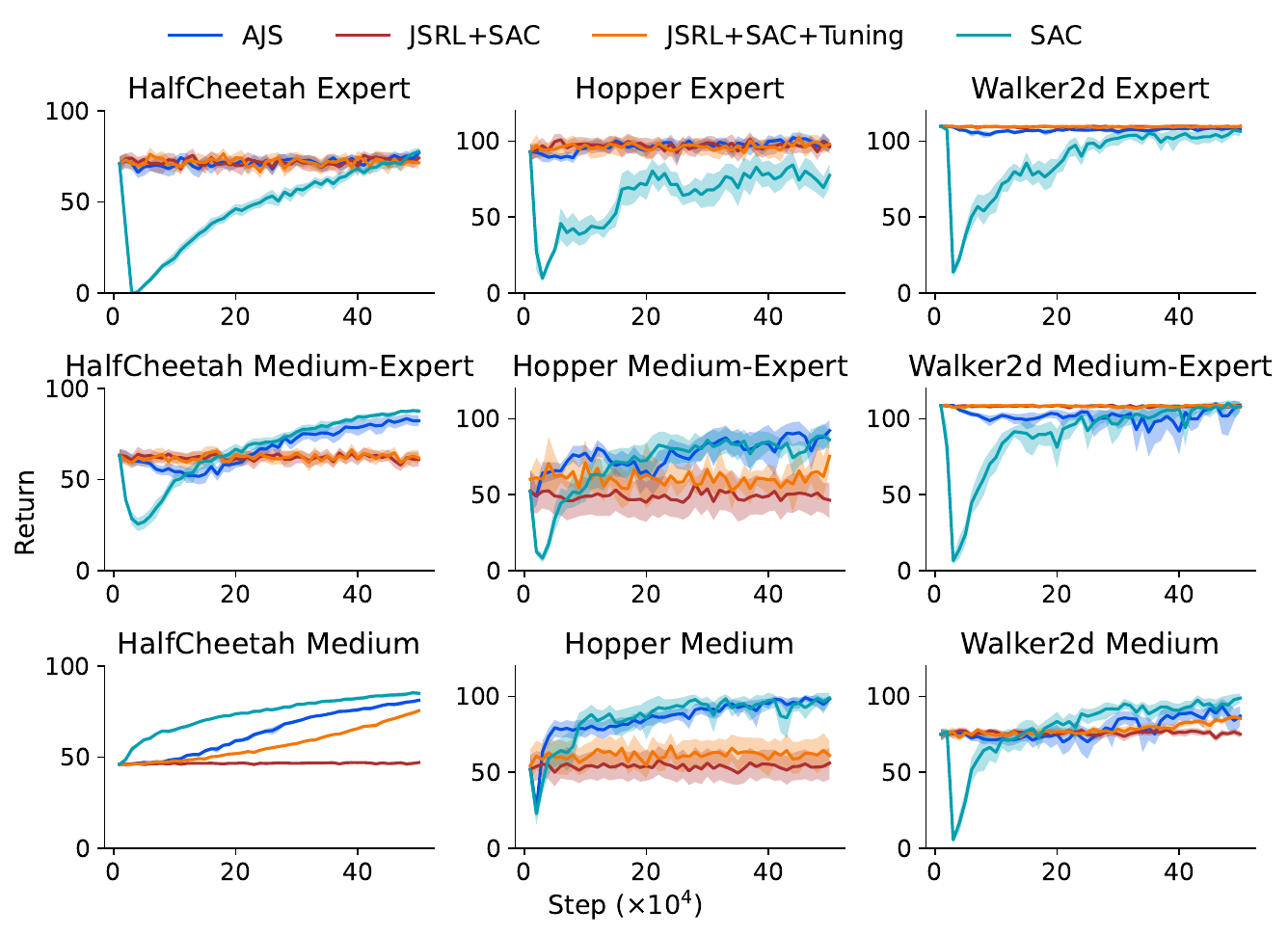}
    \vspace{-0.3cm}
    \caption{\label{fig:ajs_tunedjs}{Tuning tolerance for JSRL+SAC increased the performance improvement, while it remained no better than AJS in most cases. The x-axis is the time step. The y-axis is the normalized return. Different columns include the learning curves in HalfCheetach, Hopper, and Walker2D, respectively. The first row contains the result of the policy initialization learned from the Expert dataset. The second and third rows are initializations learned from Medium-Expert and Medium datasets, respectively. The shaded area refers to $95\%$ bootstrap confidence interval.
        }}
\end{figure}

We provide the learning curve of AJS in Figure \ref{fig:ajs_curve}. Learning curves to compare AJS and the tuned JSRL (SAC-based) are added to Figure \ref{fig:ajs_tunedjs}.

\begin{figure}[th]
    \centering
    \includegraphics[width=1\textwidth]{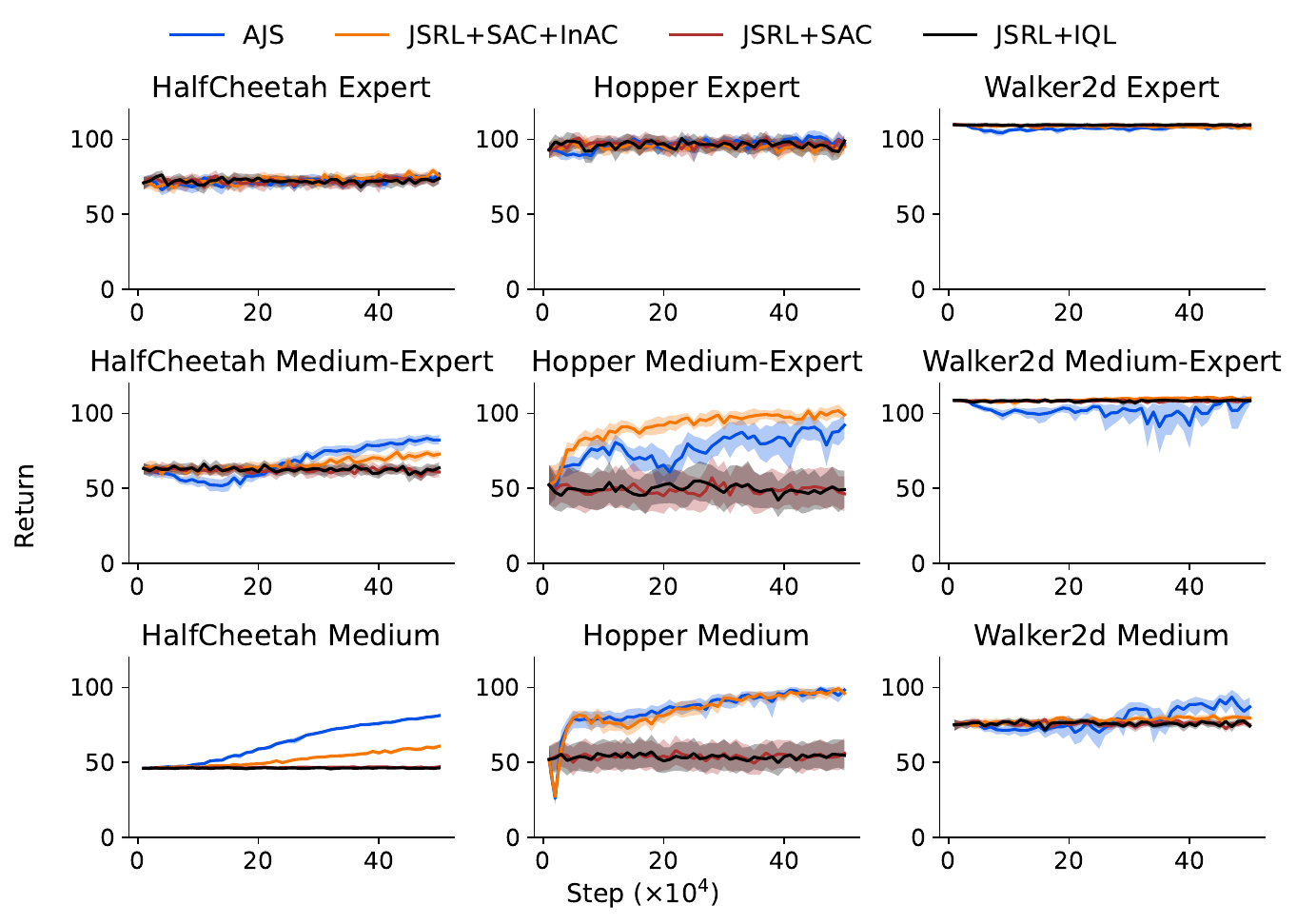}
    \vspace{-0.3cm}
    \caption{\label{fig:ajs_variants_curve}{AJS demonstrated stronger performance improvement than JSRL variants while maintaining stability. The x-axis is the time step. The y-axis is the normalized return. Different columns include the learning curves in HalfCheetach, Hopper, and Walker2D, respectively. The first row contains the result given the policy initialization learned from the Expert dataset. The second and third rows are initializations learned from Medium-Expert and Medium datasets, respectively. The shaded area refers to $95\%$ bootstrap confidence interval.
        }}
\end{figure}

\section{Experiment Details} \label{apdx:details}

\subsection{Visualizations}
Figure \ref{fig:oe_d4rl} visualizes the values at initialization of the actions before and after fine-tuning. We took the Expert dataset used for offline learning and randomly selected 1000 states. We used Principal Component Analysis (PCA) to reduce the dimensions to two for visualization~\citep{wold1987principal}. Before fine-tuning starts, we saved a copy of critic network initialization $q_{\theta,0}$, and sampled actions with the actor initialization for each of the 1000 states, written as $a_0 \sim \pi_{\phi, 0}(\cdot|s)$. At the initialization stage, the learned policy was near-optimal (Figure \ref{fig:ft_behavior_policy}). After $x$ updates, we sampled actions $a_{x} \sim \pi_{\phi, x}(\cdot|s)$ with the updated actor for the same state batch. To check whether the policy update aims toward a higher value initialization, we used the saved copy of the critic to measure the difference before the value estimates of the new policy and the initialized policy. The difference $d$ can be written as$ d = q_{\theta, 0} (s, a_x) - q_{\theta, 0} (s, a_0)$. A $d>0$ suggests that the extracted policy is not the one with the highest estimate, even the offline learning curve has been converging (Figure \ref{fig:ft_behavior_policy}).

\subsection{Policy Learning}
In all experiments, we used 2 hidden layers neural network, with 256 nodes on each layer. The batch size was 256. The online buffer size was initialized to the offline dataset size to get access to all data for offline training.

In offline learning, we fixed the learning rate to $3e-4$ and swept the temperature $\tau$. For IQL, we tried $\tau \in \{\frac{1}{3}, 0.1\}$, and also swept the expectile in the loss function in $\{0.7, 0.9\}$. For InAC, we tried $\tau \in {0.33, 0.1, 0.01}$ in Medium-Expert and Medium datasets, and used fixed $\tau=0.01$ for Expert datasets. For each setting, we run 5 random seeds. We checked the averaged final performance of the policy learned by each setting, then added 15 different seeds for the best setting. The extra 15 seeds were used for fine-tuning. In ensemble critic experiments, we used the minimum estimation of all critic networks. When updating the critic network, we used a shared target.

In online learning, the learning rate was maintained the same as in offline learning ($3e-4$). We used automatic temperature tuning for SAC and PEX \citep{haarnoja2018sacapplications}. For PROTO fine-tuning, we used automatic entropy learning as in SAC, and used the same linear update and initialization for the weight as reported in the original paper. In the IQL and AWAC experiments, we chose the same parameters as reported in the original IQL paper. JSRL and JSRL variants followed the same window size as the JSRL paper. The tolerance was set to 0 except in JSRL+SAC+Tuning, which reports the best result after tuning in $\{0\%, 5\%, 10\%\}$.


\end{document}